\documentclass[notitlepage,12pt]{article}
\usepackage[utf8]{inputenc}
\usepackage{hyperref}
\hypersetup{
    colorlinks=true,
    linkcolor=blue,
    anchorcolor=black,
    citecolor=blue,
    filecolor=cyan,
    menucolor=red,
    runcolor=cyan,
    urlcolor=blue
    }
\usepackage{authblk}
\usepackage{float}
\usepackage{verbatim}
\usepackage{bm}
\usepackage{amsmath,amsthm,amsfonts,amssymb} %
\usepackage{graphicx}
\usepackage{footnote}
\usepackage{diagbox}
\usepackage[table]{xcolor}
\usepackage{lscape}
\usepackage{longtable}
\usepackage{booktabs}
\usepackage[ruled,linesnumbered]{algorithm2e} 
\usepackage{bbm}
\usepackage[figuresright]{rotating}
\usepackage[toc, titletoc]{appendix} 
\usepackage[top=25mm, bottom=25mm, left=25mm, right=25mm]{geometry}

\newcommand{\keywords}[1]{\textbf{Keywords:} #1}
\newcommand{\U}{$\mathcal{U}$}
\newcommand{\logU}{$\log\mathcal{U}$}
\newcommand*\mean[1]{\bar{#1}}
\DeclareMathOperator{\E}{\mathbb{E}}

\DeclareMathOperator*{\argmin}{arg\,min}
\renewcommand{\Pr}{\qopname \relax m{Pr}}

\makesavenoteenv{tabular}
\makesavenoteenv{table}
\newcolumntype{?}{!{\vrule width 3pt}}

\title{Benchmarking state-of-the-art gradient boosting algorithms for classification}
\author[1*]{Piotr Florek}
\author[1]{Adam Zagdański}
\affil[1]{Faculty of Pure and Applied Mathematics, Wrocław University of Science and Technology} 
\affil[1*]{Corresponding author e-mail: \href{piotr.florek33@gmail.com}{piotr.florek33@gmail.com}}
\date{}    

\begin{document}
\maketitle

\begin{abstract}
    This work  explores the use of gradient boosting in the context of classification.
	Four popular implementations, including original GBM algorithm and selected state-of-the-art gradient boosting frameworks (i.e. XGBoost, LightGBM and CatBoost), have been thoroughly compared on several publicly available real-world datasets of sufficient diversity. In the study, special emphasis was placed on hyperparameter optimization, specifically comparing two tuning strategies, i.e. randomized search and Bayesian optimization using the  Tree-stuctured Parzen Estimator. The performance of considered methods was investigated in terms of common classification accuracy metrics as well as runtime and tuning time.
	Additionally, obtained results have been validated using appropriate statistical testing. 
	An attempt was made to indicate a gradient boosting variant showing the right balance between effectiveness, reliability and ease of use.
\end{abstract}

\keywords{gradient boosting, decision tree, classification, hyperparameter tuning}
\newpage
\section{Introduction}\label{section:intro}

Classification still remains one of the most popular Machine Learning (ML) tasks with as diverse  applications as medical diagnostics, churn analysis, fraud detection or handwritten text recognition (see e.g.\  \cite{data_classification_book} for more examples and a comprehensive overview).
Despite a large amount of research already done there is a constant need for efficient and universal algorithms that could be successfully used in a variety of scenarios, including both binary and multi-class classification problems, large data dimensionality, mixed type of attributes,  highly imbalanced or sparse datasets, etc. Moreover, reliability and ease of use appear to be equally important requirements that could make a given method the preferred choice of practitioners.

Gradient boosting is currently considered one of the most successful ML frameworks for solving both regression and classification tasks. The original formulation of the method, commonly known as Gradient Boosting Machine (GBM), was introduced by Friedman \cite{friedman_gbm,friedman_stoch} over two decades ago,  however the enormous progress in computational power and efficiency we have recently witnessed  has led to a development of new state-of-the-art gradient boosting variants. This includes but is not limited to such frameworks as eXtreme Gradient Boosting  (XGBoost) \cite{xgboost}, Light Gradient Boosting Machine (LightGBM) \cite{lightgbm} or Categorical Boosting (CatBoost) \cite{catboost}.

The effectiveness of different variants of gradient boosting in solving both business and research problems has already been widely confirmed, covering as diverse applications as e.g.\  diagnostic classification of cancers \cite{Baoshan2020}, modeling the energy consumption \cite{Touzani2018},
agricultural water management \cite{Fan2019} or customer loyalty prediction \cite{Machado2019}.
In addition, these methods often outperform competing approaches both in scientific studies dedicated to performance  benchmarking (e.g.\  \cite{Zhang2017}) as well as in various ML competitions (see e.g.\  online data science platform Kaggle\footnote{\url{kaggle.com}}). 
Unfortunately, gradient boosting algorithms tend to be highly complex and thus require good knowledge and experience on the part of the user in order to use them properly and fully realize their potential.  In particular, appropriate hyperparameter optimization or tuning is often needed to avoid suboptimal solution and speed up the training process. Moreover, with the growing number of gradient boosting variants, another practical problem arises, which is the choice of the most appropriate version of the algorithm. 

Motivated by the above limitations and challenges, in this paper  we will explore the use of gradient boosting in the context of classification, comparing in detail original GBM algorithm along with  three newer variants, i.e.: XGBoost, LightGBM and CatBoost.
The comparative analysis will be carried out on twelve publicly available datasets of sufficient diversity; in addition, special emphasis will be put on hyperparameter optimization. Specifically, the performance of GBM, XGBoost, LightGBM and CatBoost will be compared in two consecutive steps. Firstly, baseline algorithms (with no hyperparameter tuning) will be evaluated. The procedure will then be repeated with the addition of hyperparameter tuning via Bayesian optimization \cite{tpe} or randomized search \cite{randomized}. In order to avoid spurious findings, in each case, the significance of differences in the performance of the methods will be investigated using appropriate statistical methodology.

Our work can also be seen as a continuation and extension of recently published  comparative analysis \cite{comparative_analysis,comparison_of,competitive_analysis} where selected variants of gradient boosting were examined. These previous studies will be discussed in more detail in Section~\ref{section:literature}, and here we only briefly highlight the main novelty aspects in our analysis, which are threefold.
Firstly, to better reflect different real-world scenarios we decided to use more diverse benchmark data sets, including $p \gg n$ and sparse data cases. The analysis also covered unstructured data, i.e. selected image and text datasets were used. Secondly, a detailed study of influence of hyperparameter tuning on the effectiveness of methods was carried out too. While earlier studies typically used a single simplified strategy, we compared two hyperparameter optimization procedures, i.e., simple randomized search and Bayesian optimization. Finally, when benchmarking state-of-the-art gradient boosting algorithms, we tried not to limit ourselves to comparing  only the performance of the methods as expressed in terms of classification accuracy metrics and computational time.
Therefore, an attempt was made to indicate a gradient boosting variant showing the right balance between effectiveness, reliability and ease of use. 

The paper is structured as follows. In Section~\ref{section:literature} we  briefly overview related work where selected variants of gradient boosting have already been compared. Section~\ref{section:methods} introduces the methods used in this study, including short description of consecutive gradient boosting algorithms, Bayesian optimization strategy for hyperparameter tuning as well as statistical tests used for comparing classifiers.
Section \ref{section:experiments} contains the results of conducted experiments along with their brief interpretation. Section~\ref{section:discussion} is devoted to detailed discussion of results obtained, and here we also summarize our findings and recommendations in the form of a synthetic diagram. Finally, in Section~\ref{section:conclusions} the most important conclusions 
are highlighted and possible directions of future research are briefly pointed out.

\section{State of the art}\label{section:literature}

In this section, we will briefly discuss related work that has already compared several state-of-the-art gradient boosting implementations.
To begin with, let us note that performance evaluation is often an integral part of the studies proposing  a new variant of the algorithm. In particular, in \cite{xgboost} authors of XGBoost compared the performance of their algorithm with already existing frameworks, including implementation available in \emph{R} \cite{R} and \emph{scikit-learn} \cite{sklearn}. It has been concluded that in classification and learning to rank tasks XGBoost significantly outperforms basic GBM both in terms of AUC score and runtime.

In \cite{lightgbm} a thorough comparison of XGBoost and LightGBM has been performed. The datasets used contained millions of samples allowing for  detailed scalability testing. Two variants of XGBoost (i.e. exact and histogram algorithm) and three variations of LightGBM have been used, however, it has been shown that LightGBM with GOSS and EFB (i.e. with two proprietary algorithms) was the fastest and achieved the highest value of AUC score both on sparse and dense datasets. In case of two biggest datasets, LightGBM was faster than XGBoost more than 37 and 15 times, respectively. The results obtained by the creators of LightGBM clearly indicate that using GOSS and EFB simultaneously will greatly reduce the fitting time of the algorithm without sacrificing the accuracy or AUC score.

A comparison of CatBoost, XGBoost and LightGBM has been carried out in \cite{catboost}. Authors have considered both baseline and tuned models using hyperparameter search, with hyperparameter tuning performed using Bayesian optimization with Tree-structured Parzen Estimators. It has been concluded that across all datasets, CatBoost yielded the most prominent results
, although XGBoost and LightGBM performed only slightly worse. Interestingly, a baseline version of CatBoost (the one which is very similar to the original gradient boosting implementation in \cite{friedman_gbm}) without hyperparameter optimization performed better than tuned versions of XGBoost and LightGBM.

Unfortunately, a potential shortcoming of the aforementioned studies \cite{xgboost, lightgbm,catboost} may be their subjectivity, i.e., the authors of XGBoost, LightGBM or CatBoost, respectively, may have chosen such experimental scenarios and/or hyperparameter settings that favor a particular gradient boosting implementation. Therefore, it is crucial to take into the consideration comparative analyses conducted by impartial researches. Recently, the results of several such studies have been published.

In \cite{competitive_analysis} four gradient boosting variants  have been compared, including: XGBoost, LightGBM, CatBoost and SnapBoost. 
Similarly as in \cite{catboost}, both baseline and tuned (using Tree-structured Parzen Estimators) models have been compared on four diverse datasets. It has been concluded that XGBoost and SnapBoost were the most consistent in terms of accuracy and fitting time across all four datasets. Also, CatBoost was the most accurate in the case of two datasets, although it was the slowest across three. 
LightGBM and SnapBoost needed the least amount of time to be trained. Moreover, it is noticed that hyperparameter tuning improved models' performance significantly. Overall, the conclusions indicate that 
there is no single  method  that would uniformly outperforms competing  gradient boosting variants in terms of accuracy and computational time.

A significantly more detailed analysis has been carried out in \cite{comparative_analysis}, where Random Forest, GBM, XGBoost, GOSS LightGBM and Ordered CatBoost have been carefully compared.  An emphasis has been put on new specific characteristics of algorithms with respect to original GBM, i.e.: the minimum loss reduction ($\gamma$) term in XGBoost, Gradient-based One-Side Sampling (GOSS) in LightGBM and a permutation-based approach which deals with the prediction shift in CatBoost. In the study a variety across 28 used datasets was substantial including numerical, categorical, sparse and dense datasets. The diversity was also reflected in data sizes and dimensions as well as in the number of classes (varied from 2 to 18).
The authors have approached  the optimization of hyperparameters  using standard grid search, and they have also explicitly  recommended the choice of important hyperparameters for tuning. 

In \cite{comparative_analysis} it is observed that Ordered CatBoost tuning took the longest despite having the smallest defined search space. Moreover, Ordered CatBoost provided very competitive results without hyperparameter tuning, which suggests that CatBoost might be the best gradient boosting implementation "out of the box" --- both tuned and non-tuned versions yielded very similar results. Also, it is mentioned that LightGBM sometimes performs the best, however, the behaviour is not consistent. On the other hand, base versions of XGBoost and GBM perform generally worse than their tuned counterparts. However, the results presented in \cite{comparative_analysis} were not always statistically significant. 
Once again, it was concluded that across all datasets there is no such algorithm that performs the best in terms of accuracy or AUC score and fitting or tuning time.

In \cite{comparison_of} three gradient boosting implementations: XGBoost, LightGBM and CatBoost have been compared in the context of classification and regression problems. Computations for the analysis have been performed on a personal computer, to take into account that quite a few practitioners do not have access to more powerful machines (which were used in~\cite{xgboost, lightgbm,catboost,comparative_analysis,competitive_analysis}). In addition to  comparing accuracy and runtime of the algorithms, authors have also taken reliability and ease of use into consideration. It has been concluded that LightGBM performs the best in terms of accuracy and fitting time.  However, it is claimed that all three models: XGBoost, LightGBM and CatBoost still managed to reach state-of-the-art performance level. 

\section{Methods}\label{section:methods}
\subsection{Gradient boosting variants}

\subsubsection{Gradient Boosting Machine (GBM)}
The original gradient boosting method (also known as Gradient Boosting Machine or GBM) was proposed by Friedman in \cite{friedman_gbm, friedman_stoch} and serves as the baseline algorithm for XGBoost \cite{xgboost}, LightGBM \cite{lightgbm} and CatBoost \cite{catboost}. GBM combines the idea of sequentially fitting weak learners with the steepest descent method  which is used in minimization of an arbitrary loss function $L\left(F(\mathbf{x})\right)$ \cite{friedman_gbm}
\begin{equation}
F^\ast = \argmin_F\E_{y,\mathbf{x}}L\left(F(\mathbf{x})\right) = \argmin_F\E_\mathbf{x}\left[\E_y\left(L\left(F(\mathbf{x})\right)\right)\!\mid\!\mathbf{x}\right],
\end{equation}
where $F(\mathbf{x})$ is a function which  maps instances $\mathbf{x}$ to responses $y$.  
In order to approximate  a function $F^\ast(\cdot)$ an additive expansion is used in the form of 
\begin{equation}\label{eq:F_m}
F_m(\mathbf{x}) = F_{m-1}(\mathbf{x}) + \rho_mh(\mathbf{x};\mathbf{a}_m),
\end{equation}
where $\rho_m$ is the weight of the function $h$ --- the function $h$ itself corresponds to used weak learner and $\mathbf{a}_m$ refers to its parameters (e.g.\  split locations or splitting variables in case of regression trees). The first approximation of $F^\ast(\mathbf{x})$ can be expressed as
\begin{equation}
F_0(x) = \argmin_\rho \sum_{i=1}^N L(y_i,\rho).
\end{equation}
Friedman \cite{friedman_gbm} has proposed different variants of gradient boosting algorithm (depending on the choice of loss function $L(\cdot)$) that allow to solve different supervised learning tasks. Since our focus is on classification, we only present a gradient boosting algorithm intended for binary classification  (see Algorithm~\ref{alg:gb_3}), with the negative binomial log-likelihood loss function (log loss) and regression trees used as base learners. 

\begin{algorithm}[H]
\caption{Gradient boosted decision trees for classification}\label{alg:gb_3}
 $\displaystyle F_0(x) = \frac12\log\frac{1+\mean{y}}{1-\mean{y}}$\\
 \For{$m=1$ to $M$}{\vspace{5pt}
  $\displaystyle\tilde{y}_i = 2y_i\mathbin{/}\left(1+\exp(2y_iF_{m-1}(\mathbf{x}_i))\right),\, i=1,2,\ldots,N$\\[3pt]
  $\displaystyle \left\{R_{jm}\right\}_1^J = J\text{---terminal node} \;tree \left(\{\tilde{y}_i, \mathbf{x}_i\}_1^N\right)$\\
  $\displaystyle\gamma_{jm} = \sum_{\mathbf{x}_i\in R_{jm}}\tilde{y}_i\,\mathbin{/}\hspace{-7pt}\sum_{\mathbf{x}_i\in R_{jm}}|\tilde{y}_i|(2-|\tilde{y}_i|),\, j=1,2,\ldots,J$\\
  $\displaystyle F_m(\mathbf{x}) = F_{m-1}(\mathbf{x})+\nu\sum_{j=1}^J\gamma_{jm}\mathbbm{1}(\mathbf{x}\in R_{jm})$\vspace{5pt}
 }
\end{algorithm}

Note that in Algorithm~\ref{alg:gb_3} a regularization (or shrinkage) hyperparameter $0\le\nu\le 1$  has been introduced. It controls the learning rate of the algorithm and it has been shown by Friedman \cite{friedman_gbm} that its small values (such as $\nu\le 0.1$) result in a smaller generalization error.

After completing $M$ iterations of the algorithm, the ultimate approximation $F_M(\mathbf{x})$  can be used to estimate probabilities of belonging to either one of the classes, i.e.:
\begin{equation}
p_{+\mathbin{/}-}(\mathbf{x}) = \widehat{\Pr}(y=\pm 1\!\mid\!\mathbf{x}) = \left(1+e^{\pm 2F_M(\mathbf{x})}\right)^{-1},
\end{equation}
and the final prediction $\hat{y}(\mathbf{x})$ is given by
\begin{equation}
\hat{y}(\mathbf{x}) = 2\cdot \mathbbm{1}\left[p_+(\mathbf{x}) > p_-(\mathbf{x}) \right] - 1 \in \{-1,1\}.
\end{equation}

In Friedman \cite{friedman_stoch} it has been proven that introducing randomness in form of subsampling can lead to increase of performance over the original gradient boosting algorithm. 
It is worth mentioning that in addition to row-wise subsampling, feature-wise randomness can be also introduced\footnote{see e.g.\  \url{xgboost.readthedocs.io/en/stable/parameter.html}}. 
Gradient boosting natively does not support multi-class classification. Instead, usually the One-vs-Rest strategy is used.

\subsubsection{eXtreme Gradient Boosting (XGBoost)}
Authors of XGBoost \cite{xgboost} proposed many improvements over the original GBM algorithm, including introduction of regularized learning objective, approximate algorithm for split finding and sparse data support. Additionally, they claimed that XGBoost scales well with big amount of data due to having several computational optimizations (e.g.\  GPU training or out-of-core computation).

XGBoost uses modified regularized loss function \cite{comparative_analysis} in the form of 
\begin{equation}\label{eq:xgboost_loss}
L_{xgb} = \sum_{i=1}^N L\left(y_i, F(\mathbf{x}_i)\right) + \sum_{m=1}^M \Omega\left(h(\mathbf{x};\mathbf{a}_m)\right)
\end{equation}
where
\begin{equation}
\Omega\left(h(\mathbf{x};\mathbf{a}_m)\right) = \gamma T + \frac12\lambda \lVert w\rVert^2 = \gamma T +\frac12\lambda\sum_{j=1}^T w_j^2,
\end{equation}
where $\gamma$ is minimum loss reduction required to make a split, $T$ is the tree size (number of leaves), $w$ is a vector of leaf scores and $\lambda$ controls the strength of L2 regularization. Hyperparameter $\gamma$ is unique to XGBoost --- in each iteration, after construction of corresponding tree $\gamma$ is used to prune it; $\gamma$ and cost-complexity pruning hyperparameter used in decision trees have similar interpretations, however, $\gamma$-pruning is also dependent on the value of $\lambda$ \cite{xgboost}. On the other hand, $\lambda$ can be also used in implementations of LightGBM \cite{lightgbm} and CatBoost \cite{catboost}. Additionally, XGBoost in fact uses a second-order Taylor approximation of the loss function \cite{xgboost}, \cite{add_log}, therefore the user can specify a custom loss function --- then, the algorithm takes the gradient and hessian of aforementioned loss function as an input.

XGBoost supports several different splitting algorithms \cite{xgboost}, namely: \emph{exact} (which will be used in our study), \emph{approx}, \emph{hist} and \emph{gpu hist}.
Additionally, sparse features as well as missing values can be handled \cite{comparative_analysis}, \cite{comparison_of} --- during training process, XGBoost ignores aforementioned entries when determining the optimal split and then allocates them to either side of the split (which also speeds up the fitting time). In \cite{xgboost} it has been also mentioned that the sparsity-aware splitting algorithm greatly reduces the runtime of the algorithm compared to some other gradient boosting implementations that do not use sparse data enhancements.

\subsubsection{Light Gradient Boosting Machine (LightGBM)}
The LightGBM \cite{lightgbm} development process was motivated primarily by the limitations of XGBoost, as its scalability was found to be unsatisfactory for a large number of samples and features. The most notable improvements introduced in LightGBM are: leaf-wise tree growth, histogram-based splitting algorithm,  novel sample weighting Gradient-based One-Side Sampling (GOSS) and Exclusive Feature Bundling (EFB) which decrease the time to process all observations and features, respectively.

Leaf-wise tree growth algorithm produces trees with less regular structure (i.e. trees might have great depth but low number of leaves) and tends to achieve lower overall values of the loss function and faster training time. On the other hand, such trees are prone to overfitting especially when the number of training samples is low or when the tree grows very deeply --- both drawbacks can be alleviated by mindfully controlling the 
maximum depth of the tree. 

Histogram-based splitting algorithm has been implemented in XGBoost \cite{xgboost} and LightGBM \cite{lightgbm}. It aggregates values of a given feature into bins which greatly increases the scalability of the algorithm; it can also help to reduce overfitting.
Nonetheless, there is a trade-off between decreased runtime and model performance --- the smaller number of bins, the less time will be needed to train the model, but the more likely degradation in performance.

The idea of Gradient-based One-Side Sampling (GOSS) originates from AdaBoost \cite{adaboost}, where misclassified samples have higher probability of being included in subsequent training iterations. Using GOSS can lead to decreased training time while preserving accuracy. In \cite{lightgbm} it is assumed that instances with small absolute value of gradient are well-trained since the error for those instances is already small. Moreover, samples identified by big absolute value of gradient (i.e. the rate of change of the loss function) are deemed as more relevant because they cause higher fluctuations in the loss function, thus they should be prioritized during the training process. However, discarding values with small gradient entirely would decrease model's performance.

Exclusive Feature Bundling (EFB) algorithm has been proposed in order to deal more efficiently with datasets containing high number of features. In \cite{lightgbm} it has been mentioned that highly dimensional data is usually sparse which provides a great opportunity to design a nearly lossless procedure which bundles features into an \emph{exclusive feature bundle}. Similarly to GOSS, EFB has also been designed to speed up the training time without sacrificing the model's accuracy. 

In addition to the aforementioned properties, it is noteworthy that in LightGBM both L1 as well as L2 regularization is supported. Finally, it should also be emphasized that LightGBM provides a large number of hyperparameters that can significantly affect the effectiveness of the algorithm and need to be optimized to realize the full potential of the method.

\subsubsection{Categorical Boosting (CatBoost)}

CatBoost \cite{catboost} has been designed to effectively process categorical features during the training process as well as to address the so called  ''prediction shift'' which occurs in almost any implementation of gradient boosting and may have a significant impact on the model's performance. 

In \cite{catboost} authors of CatBoost show theoretical explanation of the prediction shift phenomena and propose an Ordered Boosting algorithm which tackles the aforementioned problem by using a permutation based approach. It has been concluded that addressing the prediction shift yields slightly better results when compared with a traditional variant of the algorithm (called Plain boosting). 

It is claimed \cite{catboost} that the prediction shift also occurs in case of the aforementioned categorical features preprocessing. Hence a novel method of handling categorical variables has been introduced too. However, in this work, the embedded categorical feature encoder will not be used  (see Section~\ref{section:datasets} for more details).

It is noteworthy that CatBoost algorithm uses symmetric regression trees as base models. On the other hand, other gradient boosting implementations, such as GBM \cite{friedman_gbm}, XGBoost \cite{xgboost} or LightGBM \cite{lightgbm} build asymmetric trees, either level-wise (GBM and XGBoost) or leaf-wise (LightGBM). Symmetric trees are balanced, they use the same split condition for each level of the tree. Two benefits of using them is decreased time needed for constructing predictions and reduced risk of overfitting \cite{catboost}. 

CatBoost by default chooses the optimal learning rate during the training process --- thus, it is likely that it is not a hyperparameter that needs to be tuned or even considered at all. On the other hand, unlike XGBoost or LightGBM, CatBoost does not handle sparse data appropriately, which might be undesirable in case of some datasets. Moreover, missing data is handled in a different way compared to XGBoost or LightGBM --- missing values are imputed with either minimum or maximum value for a given feature \cite{catboost}. Lastly, unlike XGBoost or LightGBM, CatBoost unfortunately does not support L1 regularization which may negatively affect the model's generalization ability.

\subsection{Bayesian optimization using TPE (Tree-structured Parzen Estimators)}

Bayesian optimization is an advanced alternative to both grid and randomized search procedures. Unlike randomized search, Bayesian optimization finds the optimal set of hyperparameters sequentially, i.e. in each iteration, new values for hyperparameters are chosen based on the results obtained in the previous ones \cite{tpe}. 
After a sufficient number of iterations, the algorithm is expected to find a set of hyperparameter values that are close, in terms of performance, to the optimal ones. 

The general idea of Bayesian optimization is to model the conditional probability $p(s\!\mid\!\mathbf{h})$ of performance scores $s$ given the vector of  hyperparameters $\mathbf{h}$. 
 In the case of modelling with TPE, however, Bayes rule is additionally applied. Specifically, instead of directly representing  $p(s\!\mid\!\mathbf{h})$ a procedure considers conditional density functions $p(\mathbf{h}\!\mid\!s < \alpha)$ and $p(\mathbf{h}\!\mid\! s\ge\alpha)$, where $\alpha$ is a specified quantile  which is used to divide the scores into "good" and "bad" ones \cite{hyperboost}. Then, one-dimensional Parzen windows (which are used in kernel density estimation) are utilized to model  both aforementioned distributions. 

The literature (\cite{tpe}, \cite{randomized}) does not clearly indicate which method: Bayesian optimization or randomized search is better, thus it is crucial to check their performance empirically.

\subsection{Statistical tests for comparing classifiers}

When comparing classification methods, the observed difference in performance may be too small to be statistically significant.  To resolve such concern appropriate statistical tests can be used, which directly take into account uncertainty  accompanying models comparison. 

In our study, the Wilcoxon signed-ranks test \cite{demsar} will be used in order to perform pairwise comparison of methods  over considered datasets, e.g.\  baseline and tuned version of a given gradient boosting algorithm will be compared.  Note that this is a non-parametric alternative to the standard paired t-test which cannot be used due to the lack of normality of differences in models' performance. However, our main goal is to compare four gradient boosting variants: GBM, XGBoost, LightGBM and CatBoost, thus the Wilcoxon signed-ranks test cannot be reliably used to determine the best performing model out of all four. Hence, more general procedure is needed.

 There are several approaches that can be use  to  assess the performance of multiple models across different datasets. One of them is to use the classical ANOVA test  \cite{demsar}, however it assumes normality and equal variances, which may limit its application in practice.
 A non-parametric alternative to ANOVA is the Friedman test. For each dataset, it ranks all used classifiers and in case of ties, an average rank is returned \cite{demsar}. According to the convention used in \cite{demsar}, the lower the rank the better, however, in this work a higher rank will imply better performance of the classifier.
 
 If the null hypothesis of the Friedman test is rejected (i.e. the ranks of the classifiers are different), post hoc analysis can be carried out using two popular approaches \cite{demsar}. The first one uses the Bonferroni correction and is used when the classifiers are compared to a control model. The second one is the Nemenyi test which does not specify a control model (the case which is considered in this work). For a pair of classifiers, their performances are significantly different if a difference between their corresponding average ranks exceeds the Critical Difference ($CD$) threshold. However, it should be noted that the Nemenyi test is known to be conservative \cite{demsar}, thus it can be expected that in the case of most pairwise comparisons of classifiers the difference of their average ranks may be statistically insignificant.
 
\section{Experiments}\label{section:experiments}
In this section, all four gradient boosting implementations: GBM, XGBoost, LightGBM and CatBoost have been thoroughly compared on real data. Across three conducted experiments, different methods of hyperparameter tuning have been used:
\begin{itemize}
    \item No tuning,
    \item Bayesian optimization using Tree-structured Parzen Estimators,
    \item Randomized search.
\end{itemize}
\subsection{Used datasets}\label{section:datasets}
Twelve real datasets have been taken into consideration; most of them are popular in data science community; other (\emph{prostate}, \emph{leukemia} \cite{microrray_data}, \emph{gina agnostic} and \emph{weather}) have been chosen because they have very high number of features. Some characteristics of the raw data which will be used have been presented in Table~\ref{tab:datasets}.

\begin{table}[h!]
\centering
\resizebox{380pt}{!}{%
\begin{tabular}{|c|c|c|c|c|}
\hline
\textbf{dataset}  & \textbf{\# samples} & \textbf{\# features} & \textbf{\# classes} & \textbf{type} \\ \hline
adult study\footnote{\label{kaggle}\url{kaggle.com/datasets}}       & 48842              & 13                  & 2                  & mixed         \\ \hline
heart disease\footnote{\url{archive.ics.uci.edu/ml/datasets/Heart+Disease}}     & 303                & 13                  & 2                  & mixed         \\ \hline
amazon\footref{kaggle}            & 32769              & 9                   & 2                  & categorical   \\ \hline
mushrooms\footref{kaggle}           & 8124               & 22                  & 2                  & categorical   \\ \hline
breast cancer\footref{kaggle}     & 569                & 30                  & 2                  & numerical     \\ \hline
churn\footref{kaggle}             & 7043               & 20                  & 2                  & mixed         \\ \hline
credit card fraud\footref{kaggle} & 30000              & 30                  & 2                  & numerical     \\ \hline
prostate         & 102                & 6033                & 2                  & numerical     \\ \hline
leukemia         & 72                 & 3571                & 2                  & numerical     \\ \hline
gina agnostic\footnote{\url{openml.org/search?type=data&status=active&id=1038}}     & 3468               & 970                 & 2                  & image, sparse \\ \hline
weather\footnote{\url{data.mendeley.com/datasets/4drtyfjtfy/1}}           & 1125               & 2500                & 4                  & image         \\ \hline
IMDB reviews\footref{kaggle}      & 10000              & 10000               & 2                  & NLP, sparse   \\ \hline
\end{tabular}
}
\caption{Benchmark datasets used for analysis}
\label{tab:datasets}
\end{table}

Variety across datasets summarized in \ref{tab:datasets} is substantial --- they have been chosen mindfully so that the algorithms can be tested in several different scenarios. Raw data varies in number of rows, features (both numerical and categorical), classes and also in the class imbalance and type. Some preprocessing methods have been applied to unstructured datasets: in the case of \emph{weather}, all images have been resized to 256x256 pixels and converted to grayscale. Then, an area of size 50x50 have been cropped from the center of each picture and subsequently flattened. For \emph{IMDB reviews}, punctuation, parentheses, links and stop words have been removed from each text review. Then, lemmatization has been performed. Finally, processed text samples have been passed to one of the transformers used in Bag of Words approach --- TF-IDF transformer \cite{sklearn} with fixed dictionary size equal to ten thousand words have been used. A big sparse matrix is produced, which will allow to test sparsity awareness of each gradient boosting implementation. 

Moreover, since none of the gradient boosting frameworks support categorical splits in regression trees, preprocessing of categorical variables has to be performed. It has been concluded that each categorical feature will be converted to a numerical one using an algorithm that is available in CatBoost \cite{catboost}. The encoding can be employed with any classifier without using CatBoost itself\footnote{Implementation of the encoding can be found in the \emph{category encoders} package:\\\indent \url{contrib.scikit-learn.org/category_encoders/catboost.html}}. It is worth noting that XGBoost, LightGBM and CatBoost are capable of performing categorical encodings on their own, but we want to keep the categorical data preprocessing consistent across all used models.

All experiments have been performed using \emph{Python} and Google Colab with 25GB of RAM memory available --- used packages as well as their versions have been listed in the Appendix.

\subsection{Performance metrics and model selection}
In our study, three popular evaluation metrics have been chosen, including accuracy, F1 score and AUC score. Aforementioned criteria measure different qualities of a classifier, which allows for a comprehensive evaluation and may expose some strengths and weaknesses
of considered gradient boosting variants. Additionally, we have considered the runtime of each model as well as tuning time.

Note that F1 and AUC measures are intended for binary classification. Therefore, in case of  multi-class classification  weighted F1 score and weighted AUC have been employed, accordingly. Weighted F1 score calculates the F1 score for each class, then a weighted mean of the scores is derived with weights corresponding to the number of instances from a given class. Similarly, weighted AUC score is computed. 

A mindful evaluation procedure has been designed to assess the performance of GBM, XGBoost, LightGBM and CatBoost --- two separate stratified cross-validation schemes for hyperparameter tuning and model evaluation have been used; a 5-fold tuning and 10-fold evaluation one. It is computationally expensive, however, it should evaluate model's performance in a reliable and realistic way. Also, to ensure consistency and reproducibility of results, for given dataset cross-validation splits (both tuning and evaluation) are exactly the same. For each dataset, the same samples will be used to train all four gradient boosting models in all iterations of cross-validation. 

\subsection{Comparative analysis of baseline models}\label{section:baseline}
\begin{figure}[H]
	\centering
		\scalebox{0.41}{\includegraphics{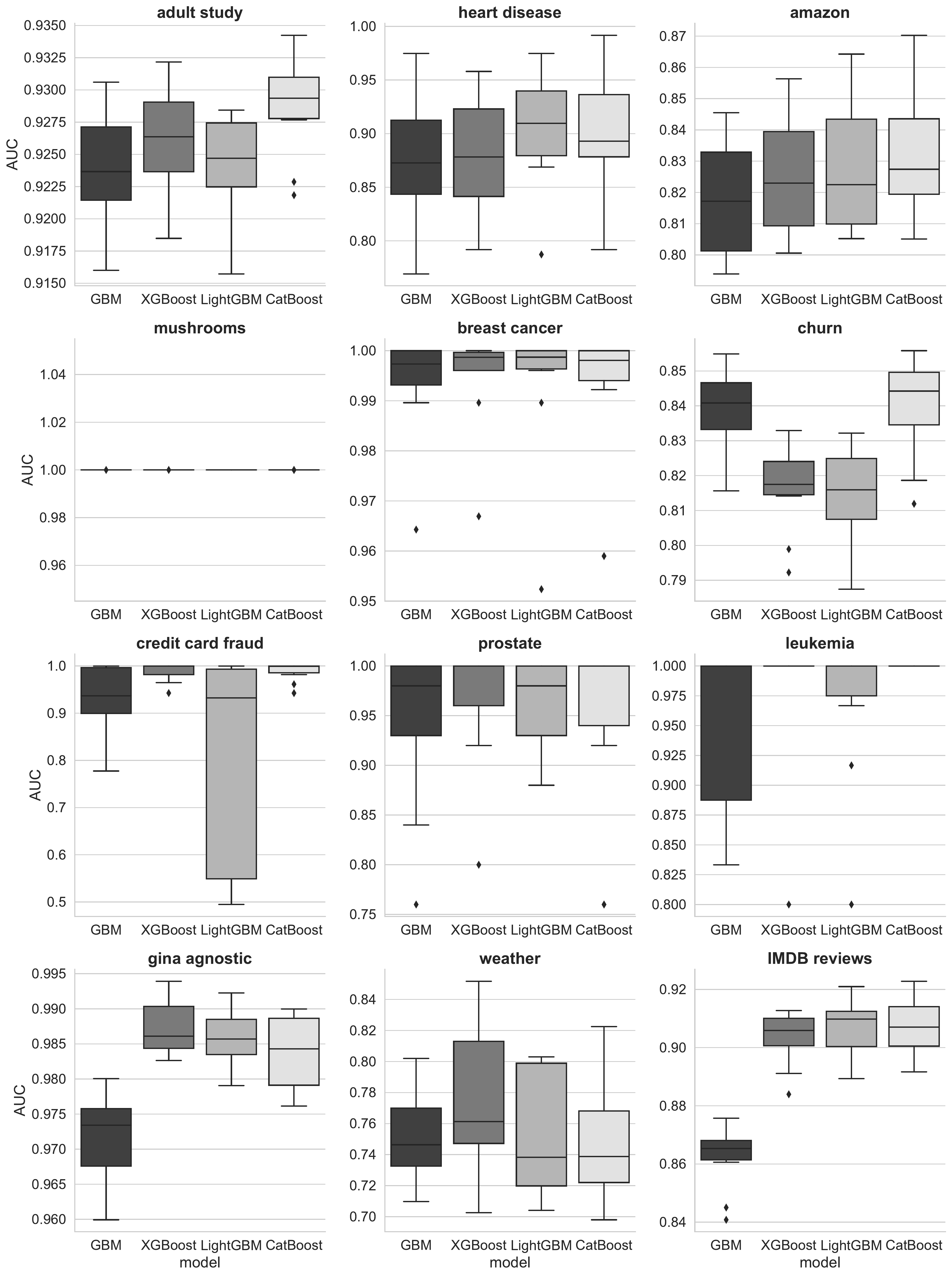}}
	\caption{AUC distributions across 12 datasets without hyperparameter tuning}
	\label{fig:no_tuning_AUC}
\end{figure}

In this experiment, baseline versions of GBM, XGBoost, LightGBM and CatBoost with minor modifications have been compared:  
\begin{enumerate}
    \item Generally, the number of trees has been set to 150 instead of the 100, which is the default value. In case of \emph{gina agnostic}, \emph{weather} and \emph{IMDB reviews} datasets  it has been set to 50 due to very high computational demands. 
    \item \emph{boosting type} in case of LightGBM has been set to GOSS.
    \item \emph{boosting type} in case of CatBoost has been set to Ordered with the exception of \emph{prostate}, \emph{leukemia}, \emph{gina agnostic}, \emph{weather} and \emph{IMDB reviews} datasets, where it has been set to Plain,  which is less computationally demanding.
\end{enumerate}
Results in terms of AUC score have been presented in Figure~\ref{fig:no_tuning_AUC}.

Performance of each of the gradient boosting implementation varies a lot when considering different datasets. There is no algorithm which would perform the best across all datasets, however, it can be observed that GBM proposed by Friedman \cite{friedman_gbm} almost always performs the worst (except for \emph{churn}, \emph{credit card fraud} and \emph{weather} datasets). Baseline versions of XGBoost and CatBoost seem to perform the best in terms of AUC score. LightGBM's performance seems unstable, especially on the \emph{credit card fraud} dataset. Results in terms of accuracy and F1 score are quite similar --- corresponding results are shown in the supplementary file. Overall, among baseline versions of GBM, XGBoost, LightGBM and CatBoost both XGBoost and CatBoost seem to perform the best it terms of accuracy, F1 score and AUC while GBM's performance is the worst.

\begin{figure}[H]
	\centering
		\scalebox{0.41}{\includegraphics{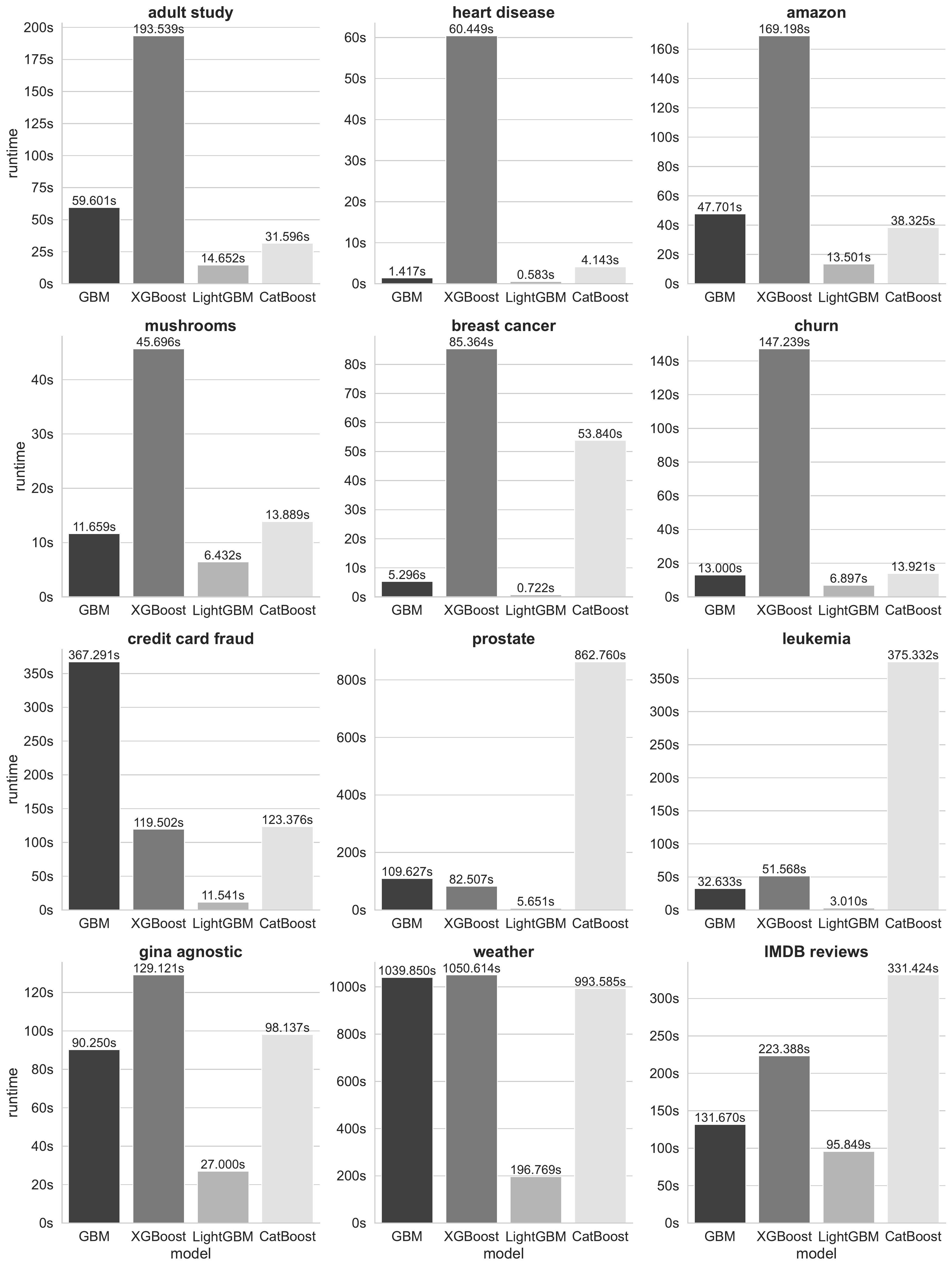}}
	\caption{Runtimes of the models across 12 datasets without hyperparameter tuning}
	\label{fig:no_tuning_runtimes}
\end{figure}

The runtimes of the algorithms are also relevant --- in case of non-tuned models, times needed to perform model evaluation have been presented in Figure~\ref{fig:no_tuning_runtimes}.

Models with the best evaluation metrics, namely XGBoost and CatBoost tend to be slower and less consistent than GBM and much slower than LightGBM which is consistently the fastest across all datasets. Standard GBM tends to perform rather quickly despite not being computationally optimized like state-of-the-art implementations.

\subsection{Models tuned with Bayesian optimization}\label{section_tpe}

In this section, comparative analysis of GBM, XGBoost, LightGBM and CatBoost models tuned using Tree-structured Parzen Estimators has been carried out. The model selection procedure is performed using a fixed number of iterations --- in our study it has been set to 15. The implementation of TPE Bayesian search can be found in \emph{tune-sklearn}\footnote{\url{github.com/ray-project/tune-sklearn}} --- an excellent package which follows the \emph{scikit-learn} API.

One of the aims of model selection in form of hyperparameter tuning is to find models with the best performance possible. Thus, it is absolutely essential to define a set of initial hyperparameters (further referred as \emph{init}) and a search space which will be used in the tuning. To preserve consistency with the experiment described in Section~\ref{section:baseline}, the number of trees and \emph{boosting\_type} in case of LightGBM and CatBoost remain unchanged. Following the advice given in \cite{comparative_analysis} and in \cite{friedman_gbm}, the number of trees has been fixed to the highest computationally feasible value (to recall, it is equal to 150 except for \emph{gina agnostic}, \emph{weather} and \emph{IMDB reviews} datasets where it has been set to 50) and consequently \emph{learning\_rate} will be one of the tuned hyperparameters.

In case of other hyperparameters used in \emph{init} and tuning, an extensive study of each of the gradient boosting implementations' documentation has been carried out. 
Additionally, Friedman's advice regarding stochastic gradient boosting \cite{friedman_stoch} has been considered, as well as the choice of hyperparameters and search spaces which were described in \cite{catboost}, \cite{comparative_analysis} and \cite{competitive_analysis}. The conclusions from the studied documentations and literature suggest the following:
\begin{enumerate}
    \item The number of trees has been fixed, but \emph{learning\_rate} is tuned.
    \item In conjunction with the advice given in \cite{friedman_stoch} and \cite{comparative_analysis}, instances and features subsampling have been set to fixed values; it is not worth to tune them. Also, \emph{subsample} cannot be used in case of LightGBM since GOSS mode is utilized.
    \item In \cite{catboost}, \cite{comparative_analysis} and \cite{competitive_analysis} the maximum depth of a tree (or number of leaves in case of LightGBM), L1 and L2 regularization were common hyperparameters which were tuned, so in this work they will also be a part of the search spaces.
    \item Also, algorithm-specific hyperparameters are a common choice during the process of model selection, thus they will also be considered. These include: XGBoost's \emph{gamma}, LightGBM's \emph{top\_rate} and \emph{other\_rate} and CatBoost's \emph{leaf\_estimation\_iterations}. 
\end{enumerate}

Additionally, some \emph{init} hyperparameters as well as some search spaces are different in case of various datasets. For those with a small number of samples, \emph{subsample} fraction has been set to 1.
Moreover, upper bound of regularization strength in search spaces has been increased in datasets with very high number of features. Also, log loss has been chosen as the optimization criterion during tuning (i.e. the optimal set of hyperparameters will minimize the value of the log loss).

The summary of initial hyperparametrization as well as search spaces has been presented in Table~\ref{tab:init_tuning}. In search spaces, discrete values in square brackets are sampled uniformly, while continuous ones are sampled either from uniform ($\mathcal{U}$) or log-uniform ($\log\mathcal{U}$) distributions.

\begin{landscape}
\begin{table}[]
\centering
\resizebox{630pt}{!}{%
\begin{tabular}{ccccc}
\hline
\multicolumn{1}{|c|}{\textbf{datasets}} &
  \multicolumn{1}{c|}{\textbf{\begin{tabular}[c]{@{}c@{}}GBM\\ init\end{tabular}}} &
  \multicolumn{1}{c|}{\textbf{\begin{tabular}[c]{@{}c@{}}XGBoost\\ init\end{tabular}}} &
  \multicolumn{1}{c|}{\textbf{\begin{tabular}[c]{@{}c@{}}LightGBM\\ init\end{tabular}}} &
  \multicolumn{1}{c|}{\textbf{\begin{tabular}[c]{@{}c@{}}CatBoost\\ init\end{tabular}}} \\ \hline
\multicolumn{1}{|c|}{\begin{tabular}[c]{@{}c@{}}adult study\\ amazon\\ mushrooms\\ churn\\ credit card fraud\end{tabular}} &
  \multicolumn{1}{c|}{\begin{tabular}[c]{@{}c@{}}n\_estimators: 150\\ subsample: 0.75\\ max\_features: 0.6\end{tabular}} &
  \multicolumn{1}{c|}{\begin{tabular}[c]{@{}c@{}}n\_estimators: 150\\ subsample: 0.75\\ colsample\_bynode: 0.6\end{tabular}} &
  \multicolumn{1}{c|}{\begin{tabular}[c]{@{}c@{}}boosting\_type: "goss"\\ n\_estimators: 150\\ colsample\_bynode: 0.6\end{tabular}} &
  \multicolumn{1}{c|}{\begin{tabular}[c]{@{}c@{}}boosting\_type: "Ordered"\\ n\_estimators: 150\\ subsample: 0.75\\ colsample\_bylevel: 0.6\end{tabular}} \\ \hline
\multicolumn{1}{|c|}{\begin{tabular}[c]{@{}c@{}}heart disease\\ breast cancer\end{tabular}} &
  \multicolumn{1}{c|}{\begin{tabular}[c]{@{}c@{}}Same as above,\\  but with subsample = 1\end{tabular}} &
  \multicolumn{1}{c|}{\begin{tabular}[c]{@{}c@{}}Same as above,\\  but with subsample = 1\end{tabular}} &
  \multicolumn{1}{c|}{\begin{tabular}[c]{@{}c@{}}Same as above\end{tabular}} &
  \multicolumn{1}{c|}{\begin{tabular}[c]{@{}c@{}}Same as above,\\  but with subsample = 1\end{tabular}} \\ \hline
\multicolumn{1}{|c|}{\begin{tabular}[c]{@{}c@{}}prostate\\ leukemia\end{tabular}} &
  \multicolumn{1}{c|}{\begin{tabular}[c]{@{}c@{}}n\_estimators: 150\\ subsample: 1\\ max\_features: 0.4\end{tabular}} &
  \multicolumn{1}{c|}{\begin{tabular}[c]{@{}c@{}}n\_estimators: 150\\ subsample: 1\\ colsample\_bynode: 0.4\end{tabular}} &
  \multicolumn{1}{c|}{\begin{tabular}[c]{@{}c@{}}boosting\_type: "goss"\\ n\_estimators: 150\\ colsample\_bynode: 0.4\end{tabular}} &
  \multicolumn{1}{c|}{\begin{tabular}[c]{@{}c@{}}boosting\_type: "Plain"\\ n\_estimators: 150\\ subsample: 1\\ colsample\_bylevel: 0.4\end{tabular}} \\ \hline
\multicolumn{1}{|c|}{\begin{tabular}[c]{@{}c@{}}gina agnostic\\ weather\\ IMDB reviews\end{tabular}} &
  \multicolumn{1}{c|}{\begin{tabular}[c]{@{}c@{}}n\_estimators: 50\\ subsample: 0.5\\ max\_features: 0.4\end{tabular}} &
  \multicolumn{1}{c|}{\begin{tabular}[c]{@{}c@{}}n\_estimators: 50\\ subsample: 0.5\\ colsample\_bynode: 0.4\end{tabular}} &
  \multicolumn{1}{c|}{\begin{tabular}[c]{@{}c@{}}boosting\_type: "goss"\\ n\_estimators: 50\\ colsample\_bynode: 0.4\end{tabular}} &
  \multicolumn{1}{c|}{\begin{tabular}[c]{@{}c@{}}boosting\_type: "Plain"\\ n\_estimators: 50\\ subsample: 0.5\\ colsample\_bylevel: 0.4\end{tabular}} \\ \hline
 &
   &
   &
   &
   \\ \hline
\multicolumn{1}{|c|}{\textbf{datasets}} &
  \multicolumn{1}{c|}{\textbf{\begin{tabular}[c]{@{}c@{}}GBM\\ tuning\end{tabular}}} &
  \multicolumn{1}{c|}{\textbf{\begin{tabular}[c]{@{}c@{}}XGBoost\\ tuning\end{tabular}}} &
  \multicolumn{1}{c|}{\textbf{\begin{tabular}[c]{@{}c@{}}LightGBM\\ tuning\end{tabular}}} &
  \multicolumn{1}{c|}{\textbf{\begin{tabular}[c]{@{}c@{}}CatBoost\\ tuning\end{tabular}}} \\ \hline
\multicolumn{1}{|c|}{\begin{tabular}[c]{@{}c@{}}adult study\\ amazon\\ mushrooms\\ churn\\ credit card fraud\end{tabular}} &
  \multicolumn{1}{c|}{\begin{tabular}[c]{@{}c@{}}max\_depth: [2, 3, 4, 5, 8, 10]\\ learning\_rate: \logU(0.01, 0.3)\\ min\_samples\_split: [2, 5, 10]\end{tabular}} &
  \multicolumn{1}{c|}{\begin{tabular}[c]{@{}c@{}}max\_depth: [2, 3, 4, 5, 8, 10]\\ learning\_rate: \logU(0.01, 0.3)\\ gamma: \U(0, 3)\\ alpha: \U(0, 1)\\ lambda: \U(0, 3)\end{tabular}} &
  \multicolumn{1}{c|}{\begin{tabular}[c]{@{}c@{}}num\_leaves: [3, 7, 15, 31, 127]\\ learning\_rate: \logU(0.01, 0.3)\\ top\_rate: \U(0.1, 0.5)\\ other\_rate: \U(0.05, 0.2)\\ reg\_alpha: \U(0, 1)\\ reg\_lambda: \U(0, 3)\end{tabular}} &
  \multicolumn{1}{c|}{\begin{tabular}[c]{@{}c@{}}max\_depth: [2, 3, 4, 5, 8, 10]\\ leaf\_estimation\_iterations: [1, 10]\\ l2\_leaf\_reg: \U(0, 5)\end{tabular}} \\ \hline
\multicolumn{1}{|c|}{\begin{tabular}[c]{@{}c@{}}heart disease\\ breast cancer\end{tabular}} &
  \multicolumn{1}{c|}{Same as above} &
  \multicolumn{1}{c|}{Same as above} &
  \multicolumn{1}{c|}{Same as above} &
  \multicolumn{1}{c|}{Same as above} \\ \hline
\multicolumn{1}{|c|}{\begin{tabular}[c]{@{}c@{}}prostate\\ leukemia\end{tabular}} &
  \multicolumn{1}{c|}{\begin{tabular}[c]{@{}c@{}}max\_depth: [2, 3, 4, 5, 8, 10]\\ learning\_rate: \logU(0.01, 0.3)\\ min\_samples\_split: [2, 5, 10]\end{tabular}} &
  \multicolumn{1}{c|}{\begin{tabular}[c]{@{}c@{}}max\_depth: [2, 3, 4, 5, 8, 10]\\ learning\_rate: \logU(0.01, 0.3)\\ gamma: \U(0, 10)\\ alpha: \U(0, 5)\\ lambda: \U(0, 10)\end{tabular}} &
  \multicolumn{1}{c|}{\begin{tabular}[c]{@{}c@{}}num\_leaves: [3, 7, 15, 31, 127]\\ learning\_rate: \logU(0.01, 0.3)\\ top\_rate: \U(0.1, 0.5)\\ other\_rate: \U(0.05, 0.2)\\ reg\_alpha: \U(0, 5)\\ reg\_lambda: \U(0, 10)\end{tabular}} &
  \multicolumn{1}{c|}{\begin{tabular}[c]{@{}c@{}}max\_depth: [2, 3, 4, 5, 8, 10]\\ leaf\_estimation\_iterations: [1, 10]\\ l2\_leaf\_reg: \U(0, 12)\end{tabular}} \\ \hline
\multicolumn{1}{|c|}{\begin{tabular}[c]{@{}c@{}}gina agnostic\\ weather\\ IMDB reviews\end{tabular}} &
  \multicolumn{1}{c|}{Same as above} &
  \multicolumn{1}{c|}{Same as above} &
  \multicolumn{1}{c|}{Same as above} &
  \multicolumn{1}{c|}{Same as above} \\ \hline
\end{tabular}%
}
\caption{Initial hyperparameters and search spaces for GBM, XGBoost, LightGBM and CatBoost}
\label{tab:init_tuning}
\end{table}
\end{landscape}

\begin{figure}[H]
	\centering
		\scalebox{0.41}{\includegraphics{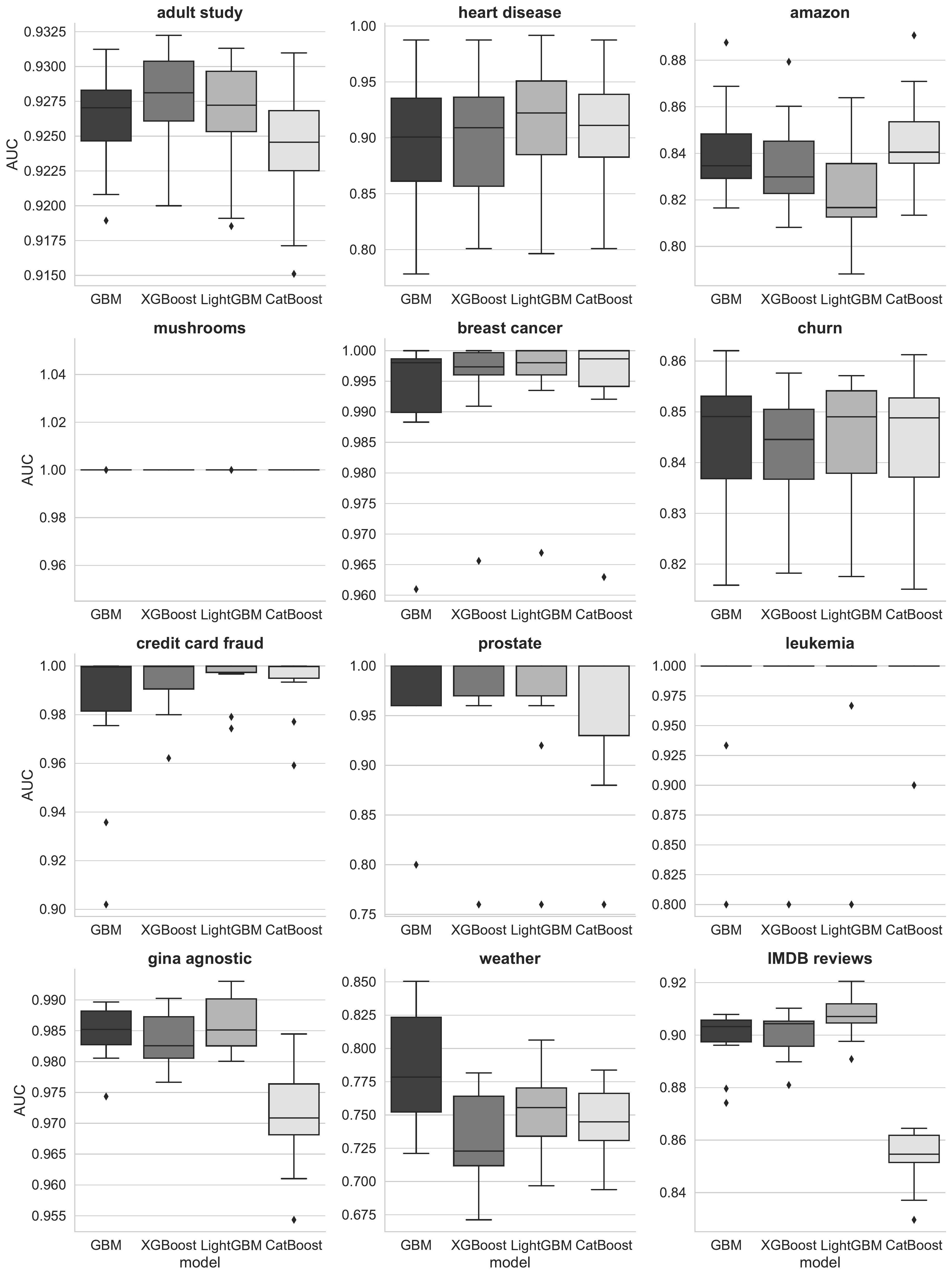}}
	\caption{AUC distributions across 12 datasets with TPE tuning}
	\label{fig:tpe_AUC}
\end{figure}

The comparison starts with the results for AUC score which have been presented in Figure~\ref{fig:tpe_AUC}.

Immediately, a vast improvement of GBM and LightGBM can be noticed. Both of the models benefited greatly from hyperparameter tuning and right now all four models: GBM, XGBoost, LightGBM and CatBoost perform very similarly, although LightGBM seems to be the best and most consistent. Identical observations can be made in the case of F1 score --- corresponding plots are shown in supplementary file.

The differences of accuracy and AUC scores between the tuned and non-tuned variants of GBM, XGBoost, LightGBM and CatBoost have been presented in Table~\ref{tab:no_tuning_tpe_accuracy_diff} and ~\ref{tab:no_tuning_tpe_AUC_diff}, respectively.
   \begin{sidewaystable}
     \centering
     \begin{tabular}{c|cccccccc}
\hline
\textbf{dataset} &
\textbf{\begin{tabular}[c]{@{}c@{}}GBM\\ mean\end{tabular}} &
\textbf{\begin{tabular}[c]{@{}c@{}}GBM\\ sdev\end{tabular}} &
\textbf{\begin{tabular}[c]{@{}c@{}}XGBoost\\ mean\end{tabular}} &
\textbf{\begin{tabular}[c]{@{}c@{}}XGBoost\\ sdev\end{tabular}} &
\textbf{\begin{tabular}[c]{@{}c@{}}LightGBM\\ mean\end{tabular}} &
\textbf{\begin{tabular}[c]{@{}c@{}}LightGBM\\ sdev\end{tabular}} &
\textbf{\begin{tabular}[c]{@{}c@{}}CatBoost\\ mean\end{tabular}} &
\textbf{\begin{tabular}[c]{@{}c@{}}CatBoost\\ sdev\end{tabular}} \\ \hline
adult study & +0.225\% & -10.035\% & +0.36\% & +38.802\% & +0.209\% & +18.611\% & -0.465\% & +18.35\%\\ \hline
heart disease & +2.893\% & -14.776\% & +2.395\% & -7.015\% & +0.805\% & +0.007\% & -0.383\% & -0.975\%\\ \hline
amazon & +0.068\% & +45.611\% & -0.2\% & +10.491\% & +0.084\% & +45.447\% & -0.052\% & +87.512\%\\ \hline
mushrooms & 0.0\% & 0.0\% & 0.0\% & 0.0\% & 0.0\% & 0.0\% & +0.012\% & -100.0\%\\ \hline
breast cancer & 0.0\% & +6.5\% & -0.901\% & -1.527\% & +0.363\% & -15.385\% & +0.181\% & +9.296\%\\ \hline
churn & +0.532\% & +32.333\% & +2.204\% & -24.691\% & +3.213\% & -46.033\% & +0.515\% & -23.365\%\\ \hline
credit card fraud & +0.037\% & -34.581\% & -0.003\% & +1.411\% & +0.291\% & -88.242\% & +0.01\% & +0.927\%\\ \hline
prostate & +3.261\% & -10.358\% & 0.0\% & 0.0\% & +6.818\% & -25.722\% & 0.0\% & 0.0\%\\ \hline
leukemia & +8.439\% & -22.171\% & -1.509\% & +1.802\% & +3.162\% & +3.705\% & 0.0\% & 0.0\%\\ \hline
gina agnostic & +2.39\% & -38.174\% & -1.276\% & +6.957\% & -0.579\% & +45.392\% & -3.277\% & +51.498\%\\ \hline
weather & +6.331\% & +54.748\% & -8.215\% & +4.31\% & +1.234\% & -29.377\% & -0.897\% & +15.341\%\\ \hline
IMDB reviews & +6.169\% & +43.014\% & -0.694\% & +21.516\% & +0.279\% & +11.814\% & -8.401\% & +38.744\%\\ \hline
\end{tabular}
\caption{Percentage gain/loss of values of means and standard deviations of accuracy --- TPE tuning vs no tuning}
\label{tab:no_tuning_tpe_accuracy_diff}

    \vspace{1.5\baselineskip}
     \centering
     \begin{tabular}{c|cccccccc}
\hline
\textbf{dataset} &
\textbf{\begin{tabular}[c]{@{}c@{}}GBM\\ mean\end{tabular}} &
\textbf{\begin{tabular}[c]{@{}c@{}}GBM\\ sdev\end{tabular}} &
\textbf{\begin{tabular}[c]{@{}c@{}}XGBoost\\ mean\end{tabular}} &
\textbf{\begin{tabular}[c]{@{}c@{}}XGBoost\\ sdev\end{tabular}} &
\textbf{\begin{tabular}[c]{@{}c@{}}LightGBM\\ mean\end{tabular}} &
\textbf{\begin{tabular}[c]{@{}c@{}}LightGBM\\ sdev\end{tabular}} &
\textbf{\begin{tabular}[c]{@{}c@{}}CatBoost\\ mean\end{tabular}} &
\textbf{\begin{tabular}[c]{@{}c@{}}CatBoost\\ sdev\end{tabular}} \\ \hline
adult study & +0.281\% & -15.696\% & +0.141\% & +3.778\% & +0.281\% & -2.272\% & -0.518\% & +23.227\%\\ \hline
heart disease & +2.207\% & -3.913\% & +2.39\% & +0.236\% & +1.201\% & +4.594\% & +0.75\% & -7.96\%\\ \hline
amazon & +2.837\% & +13.254\% & +1.226\% & +16.862\% & -0.704\% & +14.502\% & +1.609\% & +7.323\%\\ \hline
mushrooms & 0.0\% & +216.228\% & 0.0\% & -100.0\% & 0.0\% & 0.0\% & 0.0\% & -100.0\%\\ \hline
breast cancer & -0.112\% & +9.288\% & -0.027\% & +2.344\% & +0.159\% & -31.696\% & +0.053\% & -9.179\%\\ \hline
churn & +0.688\% & +18.875\% & +3.18\% & -4.343\% & +3.805\% & -8.839\% & +0.565\% & -3.227\%\\ \hline
credit card fraud & +5.143\% & -51.348\% & +0.466\% & -37.012\% & +23.897\% & -95.758\% & +0.48\% & -31.941\%\\ \hline
prostate & +2.542\% & -25.0\% & +0.415\% & +17.47\% & 0.0\% & +73.734\% & -0.833\% & +2.326\%\\ \hline
leukemia & +2.277\% & -17.382\% & 0.0\% & 0.0\% & +0.861\% & -3.154\% & -1.0\% & 0\%\\ \hline
gina agnostic & +1.333\% & -22.642\% & -0.405\% & +11.921\% & +0.014\% & +15.313\% & -1.296\% & +68.488\%\\ \hline
weather & +4.62\% & +35.32\% & -5.549\% & -23.271\% & +0.138\% & -20.273\% & -0.432\% & -23.948\%\\ \hline
IMDB reviews & +4.211\% & +9.168\% & -0.315\% & -0.261\% & +0.084\% & -7.264\% & -5.938\% & +15.504\%\\ \hline
\end{tabular}
\caption{Percentage gain/loss of values of means and standard deviations of AUC --- TPE tuning vs no tuning}
\label{tab:no_tuning_tpe_AUC_diff}
   \end{sidewaystable}

When looking at the change of mean values of accuracy and AUC one can notice that TPE tuning has increased the performance of GBM, XGBoost, LightGBM and CatBoost in terms of the mean values. The absolute value of the means in case of accuracy and AUC usually does not exceed 10\%. The most negative change of mean has been observed in the case of CatBoost's accuracy on the \emph{IMDB reviews} dataset --- TPE tuning has decreased the performance of the model by around 8.4\%. On the other hand, the biggest gain has been recorded in case of LightGBM (e.g.\  the mean AUC on the credit card fraud dataset has increased by almost 24\%). Similar conclusions can be drawn by considering F1 score instead of accuracy or AUC.

\begin{figure}[H]
	\centering
		\scalebox{0.41}{\includegraphics{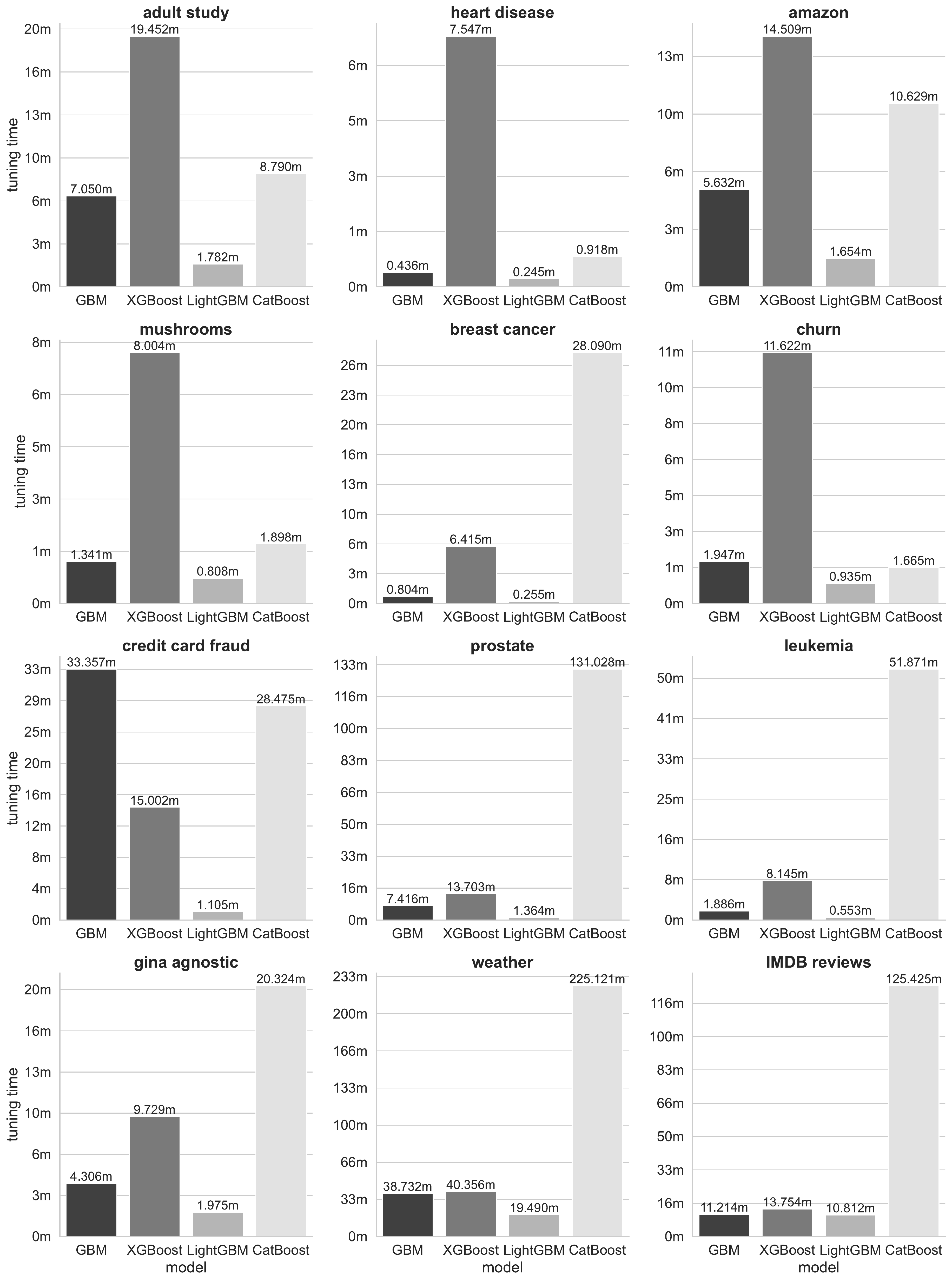}}
	\caption{Tuning times of the models across 12 datasets with TPE tuning}
	\label{fig:tpe_tuning_times}
\end{figure}

Bayesian optimization in itself is quite slow due to its sequential nature --- tuning times of the algorithms have been presented in Figure~\ref{fig:tpe_tuning_times}.

Additional time to perform model selection is quite long --- in the case of datasets with a reasonable number of features it can be stated that the time consumed to perform hyperparameter tuning is worth the slight increase of models' performance in terms of accuracy, F1 score and AUC. However, the same cannot be said in the case of highly dimensional datasets --- wait time of over two hours in the case of CatBoost will not be worth it.

\subsection{Models tuned with randomized search}\label{section_randomized}
 In this section, a different method used for model selection will be analyzed, namely the randomized search. Bayesian optimization has proven to be effective, but quite time consuming and computationally demanding, thus it is reasonable to check if simpler methods, such as aforementioned randomized search can be a better alternative.

 Each iteration of randomized search is independent from each other, so the tuning
procedure can be parallelized — thus, it is expected that 15 iterations of randomized search
will take less time than tuning using Tree-structured Parzen Estimators. Therefore, the number of
iterations has been set to 30 — results in terms of accuracy, F1 score, AUC and tuning time as well as the differences of tuning times between two aforementioned
model selection methods can be found in the supplementary file.
On most of the datasets, randomized search took longer time than TPE tuning. In
some cases the time to complete model selection was longer by over 50\%, but sometimes randomized search was faster.

It is essential to compare the performance of both types of hyperparameter tuning and the case where there was no tuning performed. The comparison has been performed for each gradient boosting implementation: GBM, XGBoost, LightGBM and CatBoost and each evaluation metric: accuracy, F1 score and AUC in the following scenarios:

\begin{enumerate}
    \item no tuning vs Bayesian optimization using TPE,
    \item no tuning vs randomized search,
    \item Bayesian optimization vs randomized search.
\end{enumerate}

All pairwise tests have been performed using the Wilcoxon signed-ranks Test with one-sided alternative hypothesis. The results have been compiled in Table~\ref{tab:no_tuning_tpe_rand}.

\begin{table}[h!]
\centering
\resizebox{450pt}{!}{%
\begin{tabular}{|c|c|c|c|c|}
\hline
\textbf{model} & \diagbox{\textbf{metric}}{\textbf{case}}         & \textbf{no tuning | TPE} & \textbf{no tuning | rand.} & \textbf{TPE | rand.} \\ \hline
               & accuracy & TPE                       & randomized                 & $p > 0.05$                 \\ \cline{2-5} 
\textbf{GBM}            & F1 score & TPE                       & randomized                 & $p > 0.05$                 \\ \cline{2-5} 
               & AUC      & TPE                       & randomized                 & $p > 0.05$                 \\ \hline
               & accuracy & $p > 0.05$                      & $p > 0.05$                       & $p > 0.05$                 \\ \cline{2-5} 
\textbf{XGBoost}        & F1 score & $p > 0.05$                      & $p > 0.05$                       & randomized           \\ \cline{2-5} 
               & AUC      & $p > 0.05$                      & $p > 0.05$                       & $p > 0.05$                 \\ \hline
               & accuracy & TPE                       & randomized                 & randomized           \\ \cline{2-5} 
\textbf{LightGBM}       & F1 score & TPE                       & randomized                 & randomized           \\ \cline{2-5} 
               & AUC      & TPE                       & randomized                 & randomized           \\ \hline
               & accuracy & $p > 0.05$                      & $p > 0.05$                       & $p > 0.05$                 \\ \cline{2-5} 
\textbf{CatBoost}       & F1 score & $p > 0.05$                      & $p > 0.05$                       & $p > 0.05$                 \\ \cline{2-5} 
               & AUC      & $p > 0.05$                      & $p > 0.05$                       & $p > 0.05$                 \\ \hline
\end{tabular}
}
\caption{Comparison of TPE and randomized search tuning in contrast to no tuning}
\label{tab:no_tuning_tpe_rand}
\end{table}

In case of GBM and LightGBM it is always more beneficial to perform either TPE tuning or randomized search rather than not. On the other hand, in case of XGBoost and CatBoost no significant differences of performing hyperparameter tuning have been found --- it may imply that XGBoost and CatBoost do not have to be tuned at all.

The final rankings of the models in terms of choice of hyperparameter tuning method and evaluation metric has been presented in Figure~\ref{tab:all_tunings_rankings}. A Friedman test with $D=12$ datasets and $k=12$ models have been performed.

\begin{table}[h!]
\centering
\resizebox{420pt}{!}{%
\begin{tabular}{|c|c|c|c?c|c|c|c|}
\hline
\textbf{\begin{tabular}[c]{@{}c@{}}model\\ rank\end{tabular}} &
  \textbf{accuracy} &
  \textbf{F1 score} &
  \textbf{AUC} &
  \textbf{\begin{tabular}[c]{@{}c@{}}model\\ rank\end{tabular}} &
  \textbf{accuracy} &
  \textbf{F1 score} &
  \textbf{AUC} \\ \hline
\textbf{\#1} &
  \begin{tabular}[c]{@{}c@{}}LightGBM\\ randomized\\ 9.75\end{tabular} &
  \begin{tabular}[c]{@{}c@{}}LightGBM\\ randomized\\ 9.62\end{tabular} &
  \begin{tabular}[c]{@{}c@{}}LightGBM\\ randomized\\ 9.46\end{tabular} &
  \textbf{\#7} &
  \begin{tabular}[c]{@{}c@{}}GBM\\Bayesian\\ 6.50\end{tabular} &
  \begin{tabular}[c]{@{}c@{}}GBM\\Bayesian\\ 6.17\end{tabular} &
  \begin{tabular}[c]{@{}c@{}}GBM\\Bayesian\\ 6.50\end{tabular} \\ \hline
\textbf{\#2} &
  \begin{tabular}[c]{@{}c@{}}LightGBM\\Bayesian\\ 7.54\end{tabular} &
  \begin{tabular}[c]{@{}c@{}}LightGBM\\Bayesian\\ 8.25\end{tabular} &
  \begin{tabular}[c]{@{}c@{}}LightGBM\\Bayesian\\ 8.38\end{tabular} &
  \textbf{\#8} &
  \begin{tabular}[c]{@{}c@{}}CatBoost\\Bayesian\\ 6.17\end{tabular} &
  \begin{tabular}[c]{@{}c@{}}CatBoost\\Bayesian\\ 6.08\end{tabular} &
  \begin{tabular}[c]{@{}c@{}}CatBoost\\Bayesian\\ 6.21\end{tabular} \\ \hline
\textbf{\#3} &
  \begin{tabular}[c]{@{}c@{}}XGBoost\\ no tuning\\ 7.50\end{tabular} &
  \begin{tabular}[c]{@{}c@{}}CatBoost\\ no tuning\\ 7.54\end{tabular} &
  \begin{tabular}[c]{@{}c@{}}XGBoost\\ randomized\\ 7.21\end{tabular} &
  \textbf{\#9} &
  \begin{tabular}[c]{@{}c@{}}CatBoost\\ randomized\\ 5.71\end{tabular} &
  \begin{tabular}[c]{@{}c@{}}CatBoost\\ randomized\\ 5.50\end{tabular} &
  \begin{tabular}[c]{@{}c@{}}GBM\\ randomized\\ 6.12\end{tabular} \\ \hline
\textbf{\#4} &
  \begin{tabular}[c]{@{}c@{}}CatBoost\\ no tuning\\ 7.12\end{tabular} &
  \begin{tabular}[c]{@{}c@{}}XGBoost\\ no tuning\\ 7.50\end{tabular} &
  \begin{tabular}[c]{@{}c@{}}XGBoost\\Bayesian\\ 6.88\end{tabular} &
  \textbf{\#10} &
  \begin{tabular}[c]{@{}c@{}}XGBoost\\Bayesian\\ 5.54\end{tabular} &
  \begin{tabular}[c]{@{}c@{}}XGBoost\\Bayesian\\ 5.33\end{tabular} &
  \begin{tabular}[c]{@{}c@{}}CatBoost\\ randomized\\ 5.96\end{tabular} \\ \hline
\textbf{\#5} &
  \begin{tabular}[c]{@{}c@{}}XGBoost\\ randomized\\ 6.88\end{tabular} &
  \begin{tabular}[c]{@{}c@{}}XGBoost\\ randomized\\ 7.04\end{tabular} &
  \begin{tabular}[c]{@{}c@{}}CatBoost\\ no tuning\\ 6.75\end{tabular} &
  \textbf{\#11} &
  \begin{tabular}[c]{@{}c@{}}LightGBM\\ no tuning\\ 5.25\end{tabular} &
  \begin{tabular}[c]{@{}c@{}}LightGBM\\ no tuning\\ 4.83\end{tabular} &
  \begin{tabular}[c]{@{}c@{}}LightGBM\\ no tuning\\ 5.21\end{tabular} \\ \hline
\textbf{\#6} &
  \begin{tabular}[c]{@{}c@{}}GBM\\ randomized\\ 6.58\end{tabular} &
  \begin{tabular}[c]{@{}c@{}}GBM\\ randomized\\ 6.62\end{tabular} &
  \begin{tabular}[c]{@{}c@{}}XGBoost\\ no tuning\\ 6.54\end{tabular} &
\textbf{\#12} &
  \begin{tabular}[c]{@{}c@{}}GBM\\ no tuning\\ 3.46\end{tabular} &
  \begin{tabular}[c]{@{}c@{}}GBM\\ no tuning\\ 3.50\end{tabular} &
  \begin{tabular}[c]{@{}c@{}}GBM\\ no tuning\\ 2.79\end{tabular} \\ \hline
\end{tabular}
}
\caption{Final rankings of 12 models for accuracy, F1 score and AUC}
\label{tab:all_tunings_rankings}
\end{table}

\begin{figure}[H]
	\centering
		\scalebox{0.47}{\includegraphics{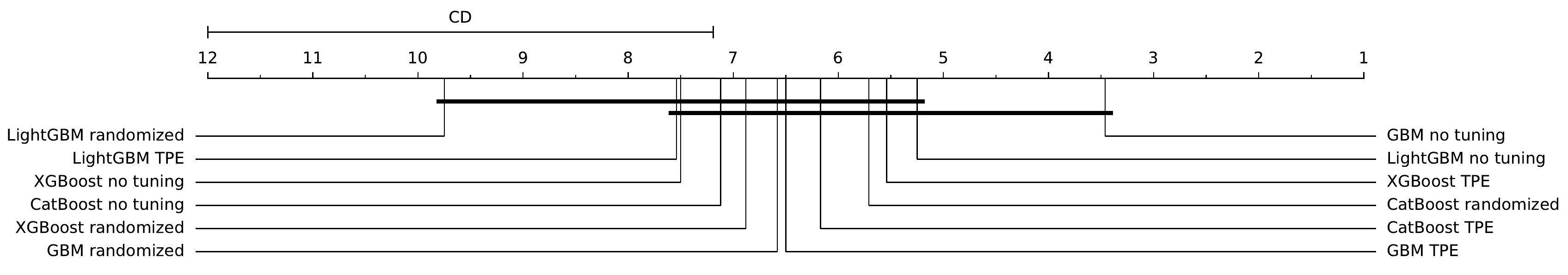}}
	\caption{Critical difference plot of the accuracy ranks from Table~\ref{tab:all_tunings_rankings} --- two bold horizontal lines connect models for which the aforementioned difference is insignificant}
	\label{fig:cd_plot}
\end{figure}

Additionally, p-values of the Nemenyi post-hoc test indicate that LightGBM tuned using randomized search is better than baseline GBM. The same is true for LightGBM tuned using TPE tuning in case of AUC score (critical difference was equal to 4.81). The ranking of all algorithms have been shown in Table~\ref{tab:all_tunings_rankings} while visualization of the ranks compared to the critical difference has been presented in Figure~\ref{fig:cd_plot}.

The visualization of ranks in Figure~\ref{fig:cd_plot} indicates that LightGBM tuned using randomized search and baseline GBM are models which performed much better and much worse than other variants, respectively. Only for these two aforementioned models the difference of their performance is statistically significant. Enormous improvement of performance in case of LightGBM and GBM compared to their non-tuned versions can be observed too. The ranking presented in Table~\ref{tab:all_tunings_rankings} confirm that in case of accuracy, F1 score and AUC tuned LightGBM performs the best by a significant margin.

\section{Discussion}\label{section:discussion}
Results presented in Sections~\ref{section:baseline}, \ref{section_tpe} and \ref{section_randomized} suggest that LightGBM after performing hyperparameter tuning is the best gradient bosting implementation according to accuracy, F1 score and AUC score while being decidedly the fastest and most efficient. Additionally, in our study, only the difference of performance of tuned LightGBM and baseline GBM was confirmed to be statistically significant. 

Authors of \cite{comparative_analysis} (see Section \ref{section:literature}) have stated that CatBoost tuned using grid search procedure was the best model in terms of accuracy --- however, its rank was statistically insignificant compared to other models. Also, in \cite{comparative_analysis} it has been stated that the performance of tuned and baseline variants of CatBoost was very similar --- on the other hand, in this work we have proven that performing hyperparameter search with CatBoost is counterproductive; its baseline variant has proven to perform much better in terms of accuracy, F1 score and AUC. In \cite{comparative_analysis} authors have mentioned that CatBoost need not to be tuned to achieve competitive performance which coincides with the results obtained in our study. 

Both in \cite{comparative_analysis} and in our analysis baseline LightGBM is one of the worst GBM variants in terms of performance. However, we have proven that baseline XGBoost is one of the best algorithms in terms of average ranks in case of F1 score and AUC (rank 3 and 4, respectively) --- in \cite{comparative_analysis} the authors have shown that baseline XGBoost performs poorly, which is surprising. Interestingly, tuned GBM turned out to be the best performing algorithm alongside tuned CatBoost and XGBoost, while in our study it has been ranked sixth or seventh out of twelve algorithms in terms of average rank. Additionally, GBM in this work was clearly the worst performing gradient boosting implementation, but in \cite{comparative_analysis} it outperformed baseline variants of XGBoost and LightGBM. The differences between the results obtained in our comparative analysis and in \cite{comparative_analysis} could be caused by the choice of different model evaluation schemes as well as benchmark datasets with different characteristics (e.g.\  $p\gg n$ case).

In \cite{competitive_analysis} CatBoost was deemed as the best accurate implementation compared to XGBoost, LightGBM and SnapBoost and it has been stated that it performed the best in case of the categorical dataset --- in our study, that was not the case for both \emph{amazon} and \emph{mushrooms} datasets. Additionally, in \cite{competitive_analysis} XGBoost, LightGBM and CatBoost all greatly benefited from hyperparameter tuning while it was not confirmed in our analysis --- even though in both cases Bayesian optimization was used, different search spaces have been provided; additionally, authors of \cite{competitive_analysis} have utilized the simple train-test split.

On the other hand, some of our findings coincide with those presented in \cite{comparison_of}. In \cite{comparison_of} it has been concluded that among tuned variants of XGBoost, LightGBM and CatBoost LightGBM performed the best both in terms of accuracy and runtime. Additionally, in \cite{comparison_of} CatBoost was the algorithm which benefited the most from hyperparamerer tuning --- overall, it was never the case that a baseline variant of XGBoost, LightGBM or CatBoost was better than their tuned counterparts; the tuning always increased the accuracy of the models. 
In terms of model fitting time, the results presented in the literature: \cite{comparative_analysis}, \cite{competitive_analysis}, \cite{comparison_of} as well as \cite{lightgbm} tend to coincide with our conclusions --- LightGBM is always the fastest algorithm, no matter if the GOSS sampling is used or not. Additionally, in this work we have shown that the runtime of CatBoost (both Plain and Ordered variants) is heavily dependent on the dimensionality of the data, but not on the number of samples. CatBoost's runtime is quite low for datasets with moderate number of features and very high in case of highly dimensional data --- such conclusion is not consistent with the literature, where the overall consensus is that CatBoost, especially its Ordered variant is the slowest (although most of the datasets which have been considered in the literature have low to moderate number of features).

Moreover, in our study we have shown that XGBoost (with the exact greedy splitting algorithm) tends to perform inconsistently in terms of runtime across different datasets. Since XGBoost offers many options regarding the splitting algorithm (i.e. exact, approximate and histogram), in practice it is reasonable to verify the performance of all three aforementioned variants.

The ease of use is also a relevant consideration when using GBM~\cite{friedman_gbm}, XGBoost~\cite{xgboost}, LightGBM~\cite{lightgbm} and CatBoost~\cite{catboost}. It was only mentioned in \cite{comparison_of}, however, in this work we came up with our own conclusions. Overall, the simplest algorithm to use is the basic GBM --- it has the least number of available hyperparameters.
We have shown that baseline variants of XGBoost and CatBoost perform very well --- they work great "out of the box", and this makes them the easiest algorithms to use. LightGBM needs to be tuned in order to achieve high performance, thus it should not be deemed as easy to use. 

In terms of the number of available hyperparameters, LightGBM offers the biggest variety. There are a lot of hyperparameters available both in XGBoost and LightGBM, but the number of possible configurations in XGBoost is not as high as in the case of LightGBM. CatBoost offers the least number of available hyperparameters, but they are often unique and cannot be found in any other gradient boosting implementation. However, CatBoost offers the least amount of information available in the documentation; on the other hand, XGBoost and LightGBM offer much more exhaustive descriptions of hyperparameters. Thus, when considering hyperparameter tuning, we have concluded that XGBoost would be the easiest to use; the second easiest GBM implementation would be LightGBM. We advise that LightGBM should be used by experienced researchers who know the ins and outs of gradient boosting algorithms and the specifics of LightGBM. However, none of GBM, XGBoost or LightGBM offer robust support of categorical and text variables --- in this context, CatBoost is the clear winner. At the time of conducting this work, XGBoost's categorical features support is  experimental and LightGBM's algorithm is not as unbiased and optimized as the one offered in CatBoost.

\begin{sidewaysfigure}
	\centering
		\scalebox{0.67}{\includegraphics{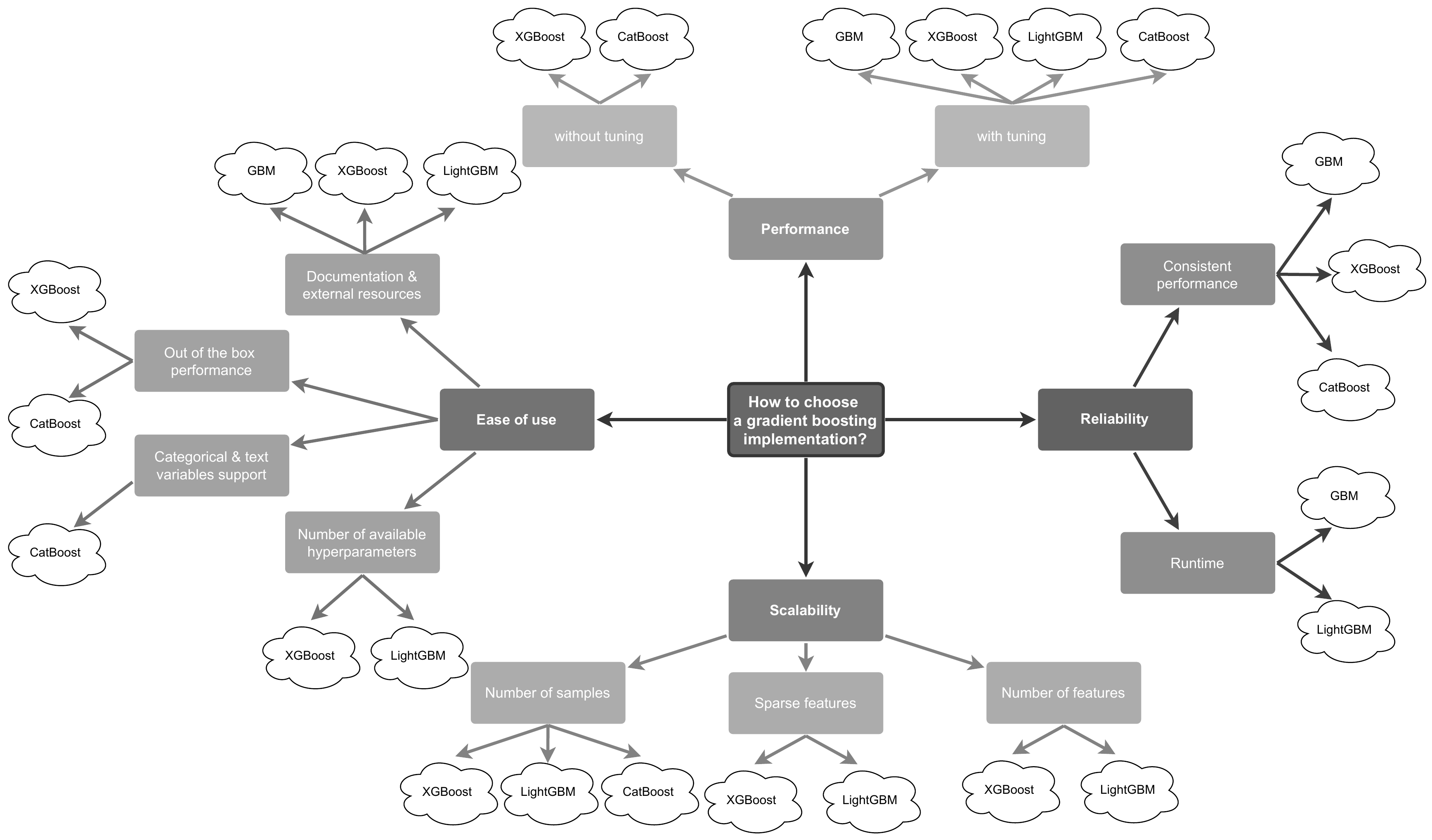}}
	\caption{The choice of optimal gradient boosting implementation in different scenarios}
	\label{fig:recommendations}
\end{sidewaysfigure}

The recommendations regarding the choice of the optimal gradient boosting variant depends on many factors --- a summary diagram has been presented in Figure~\ref{fig:recommendations}. To summarize, XGBoost and CatBoost are the best algorithms when baseline variants are considered and all models: GBM, XGBoost, LightGBM and CatBoost can be said to perform well in the case when hyperparameter tuning is performed. LightGBM was the only model with unstable performance, while GBM and LightGBM were very consistent in terms of runtime across different datasets.

In terms of scalability, due to the incorporation of L1 and L2 regularization and sparsity-aware splitting algorithms XGBoost and LightGBM should be the primary choices when considering datasets with high number of features. GBM does not scale well with the number of samples (since it hardly implements any mechanisms which may reduce the runtime) --- XGBoost, LightGBM and CatBoost seem to process large number of samples quite well. When the number of samples is very large, then LightGBM or histogram variant of XGBoost should be used --- also, GPU can be used with XGBoost, LightGBM and CatBoost and it will greatly decrease the runtimes for large datasets.

\newpage
\section{Conclusions and future research}\label{section:conclusions}
In this work, we have concluded that LightGBM tuned using either Bayesian optimization or randomized search was the best gradient boosting algorithm in terms of an average rank across twelve datasets. Due to excellent performance, very low runtime or tuning time and substantial number of useful hyperparameters tuned LightGBM has the biggest potential if used mindfully. We have
shown that all considered variants of gradient boosting perform exceptionally well, both in
terms of evaluation metrics, such as accuracy, F1 score or AUC score and runtime. However,
in most of the cases some gradient boosting implementations might be a better choice than other.
Although some general recommendations regarding choosing the best gradient boosting
algorithm can be given, each use case is highly individual --- still, the choice of an optimal gradient boosting variant remains an open
problem.

\section*{Declarations}
\begin{itemize}
    \item {\bf Data availability:} All benchmark datasets used in the study are publicly available (see Table~\ref{tab:datasets} for more details on  accessibility).
    \item {\bf Conflict of interest:} The authors declare that they have no conflict of interest.
    \item {\bf Funding:} The authors received no financial support for the research, authorship, and/or publication of this article.
\end{itemize}

\begin{appendices}\label{appendix}

\section*{Appendix}
The \emph{Python} version used in the experiments is 3.9.6.
Packages containing gradient boosting implementations, Bayesian optimization framework and categorical variables encoding algorithm are the following: 

\begin{itemize}
\item scikit-learn==1.0.2
\item xgboost==1.5.2
\item lightgbm==3.3.2
\item catboost==1.0.4
\item tune-sklearn==0.4.1
\item ray[tune]==1.10.0
\item hyperopt==0.2.7
\item category\_encoders==2.4.1
\end{itemize}

\end{appendices}

\bibliographystyle{ieeetr}
\bibliography{references.bib}

\end{document}


\maketitle

\begin{figure}[H]
	\centering
		\scalebox{0.43}{\includegraphics{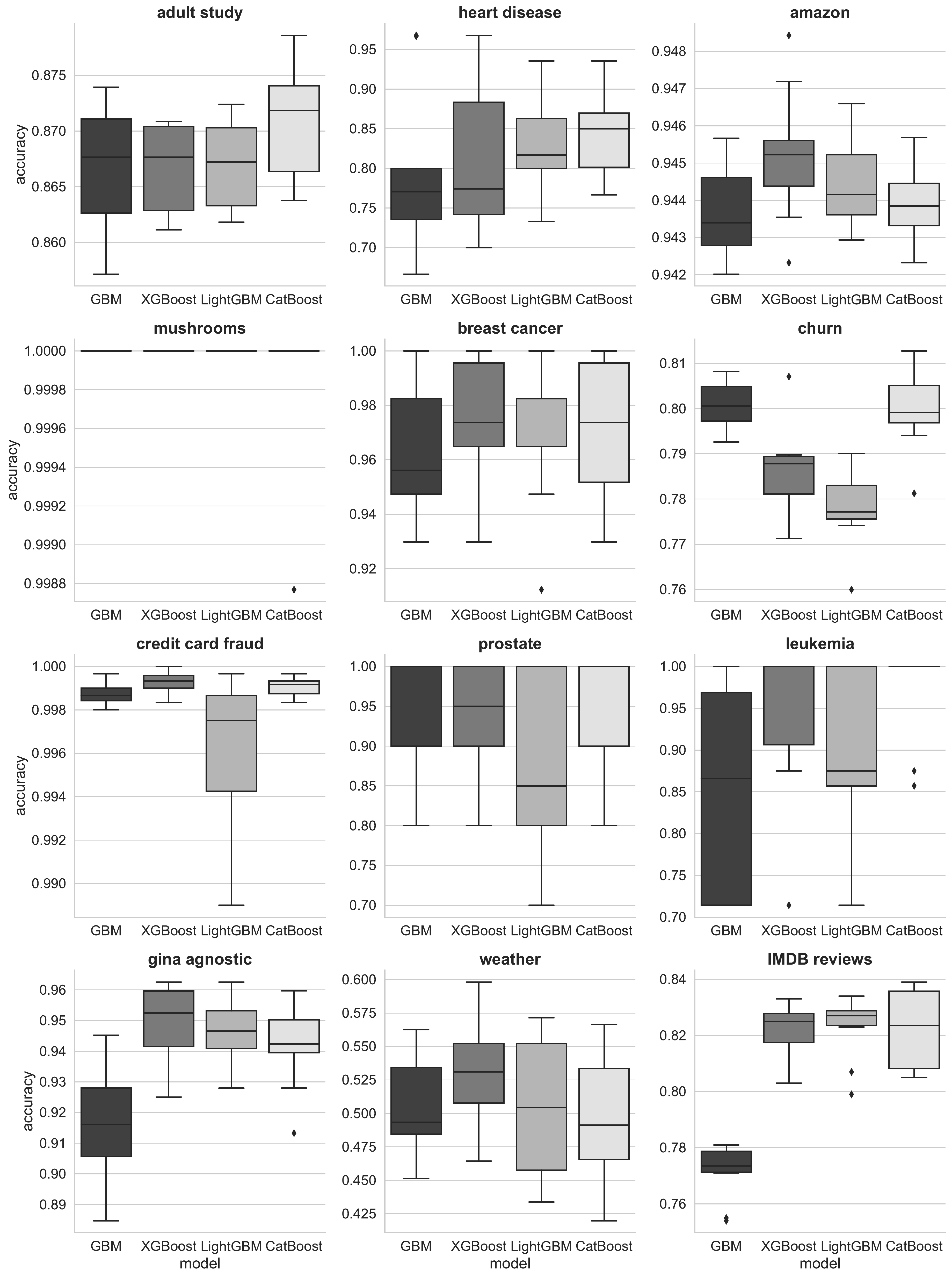}}
	\caption{Accuracy distributions across 12 datasets without hyperparameter tuning}
	\label{fig:no_tuning_accuracy}
\end{figure}

\begin{figure}[H]
	\centering
		\scalebox{0.43}{\includegraphics{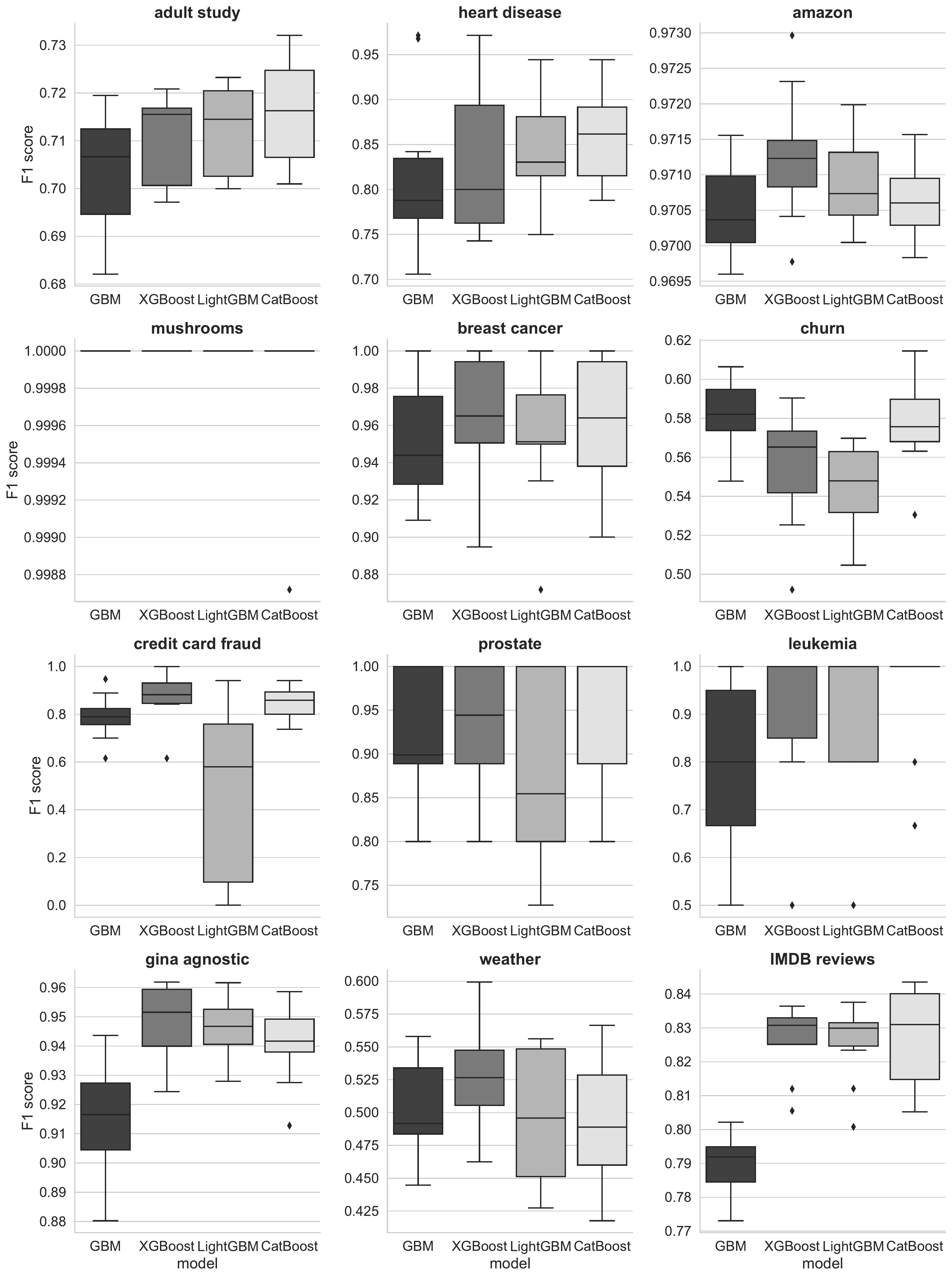}}
	\caption{F1 score distributions across 12 datasets without hyperparameter tuning}
	\label{fig:no_tuning_f1}
\end{figure}

\begin{figure}[H]
	\centering
		\scalebox{0.43}{\includegraphics{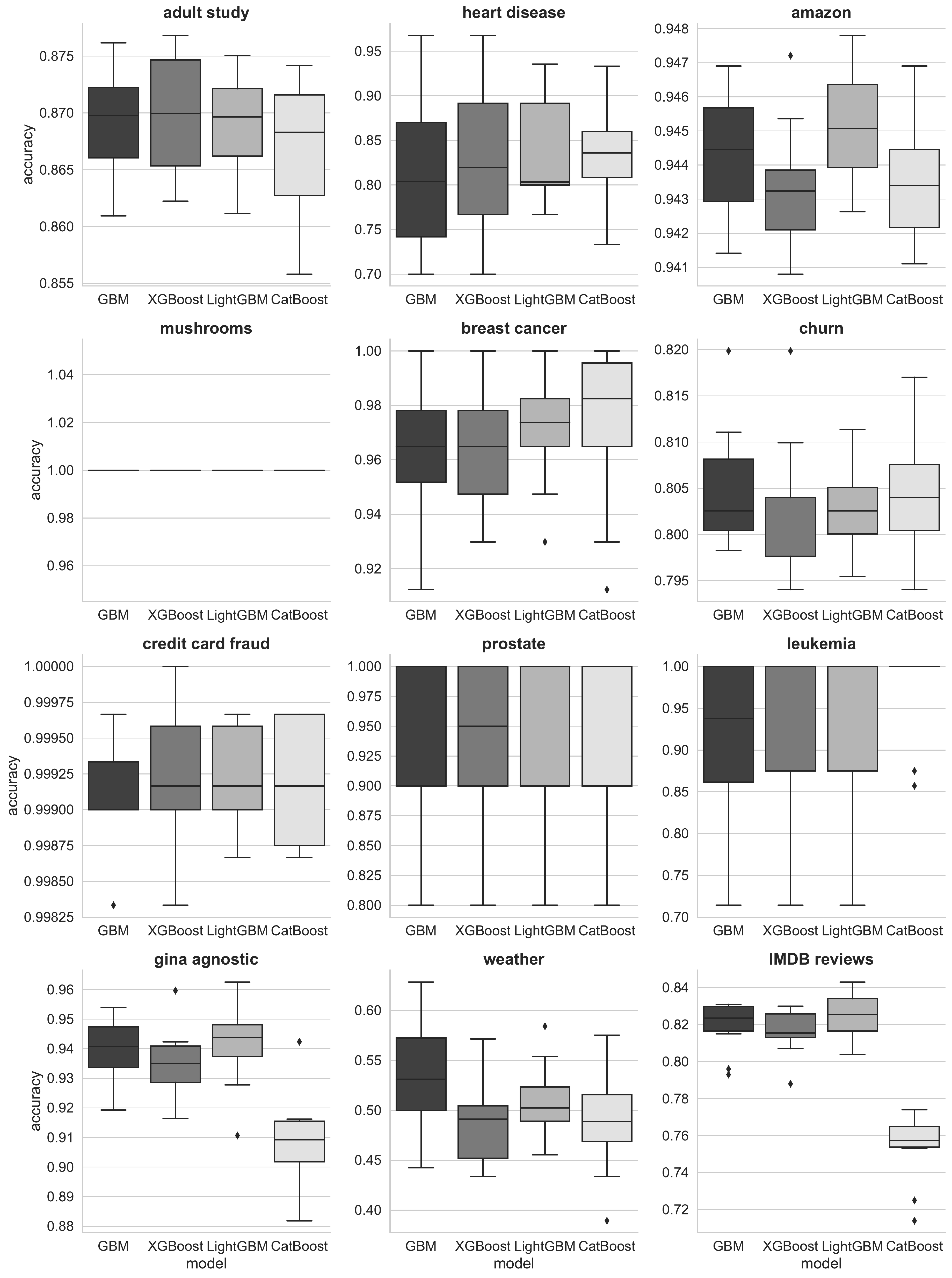}}
	\caption{Accuracy distributions across 12 datasets with TPE tuning}
	\label{fig:tpe_accuracy}
\end{figure}

\begin{figure}[H]
	\centering
		\scalebox{0.43}{\includegraphics{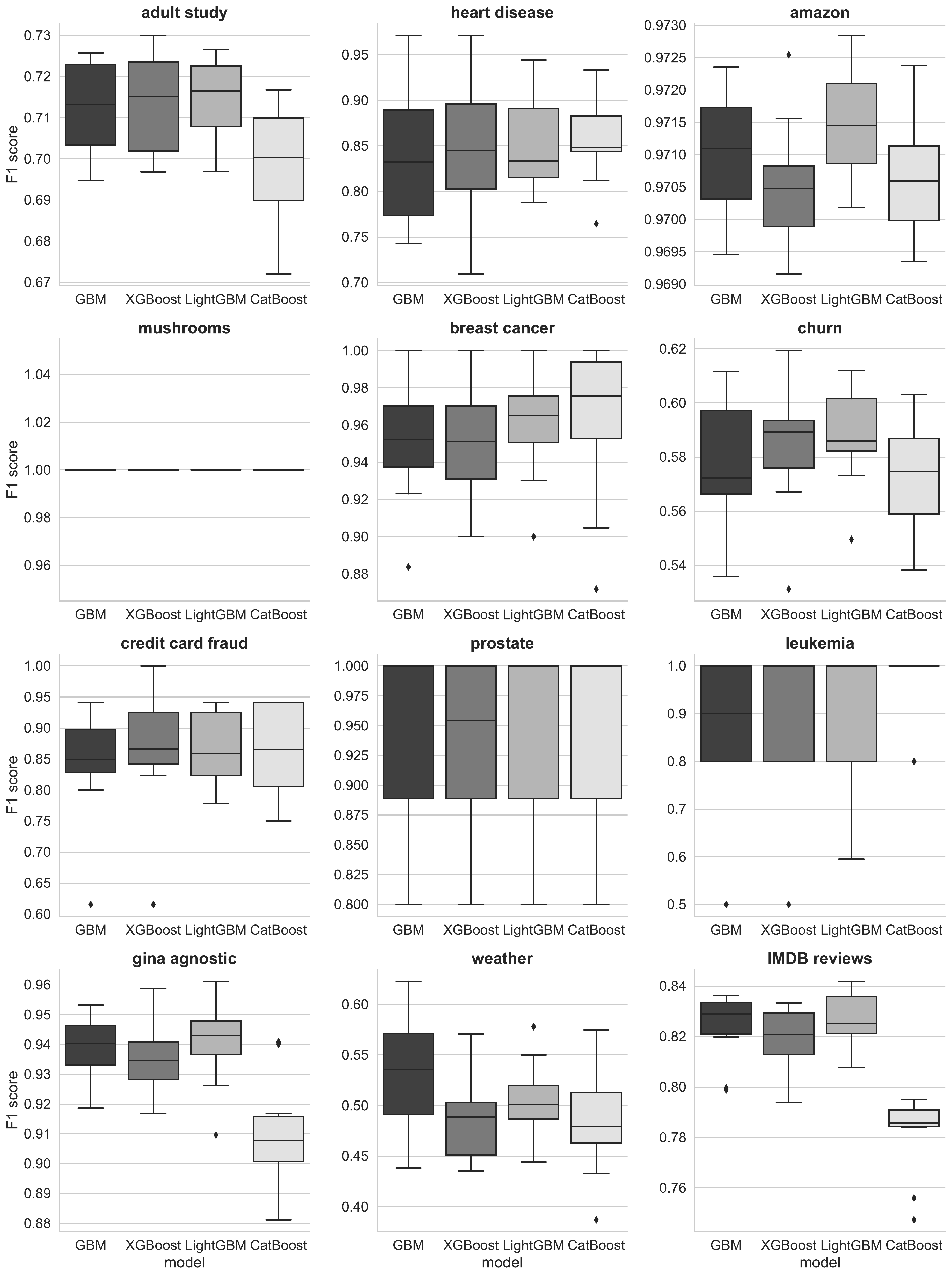}}
	\caption{F1 score distributions across 12 datasets with TPE tuning}
	\label{fig:tpe_f1}
\end{figure}

\begin{figure}[H]
	\centering
		\scalebox{0.43}{\includegraphics{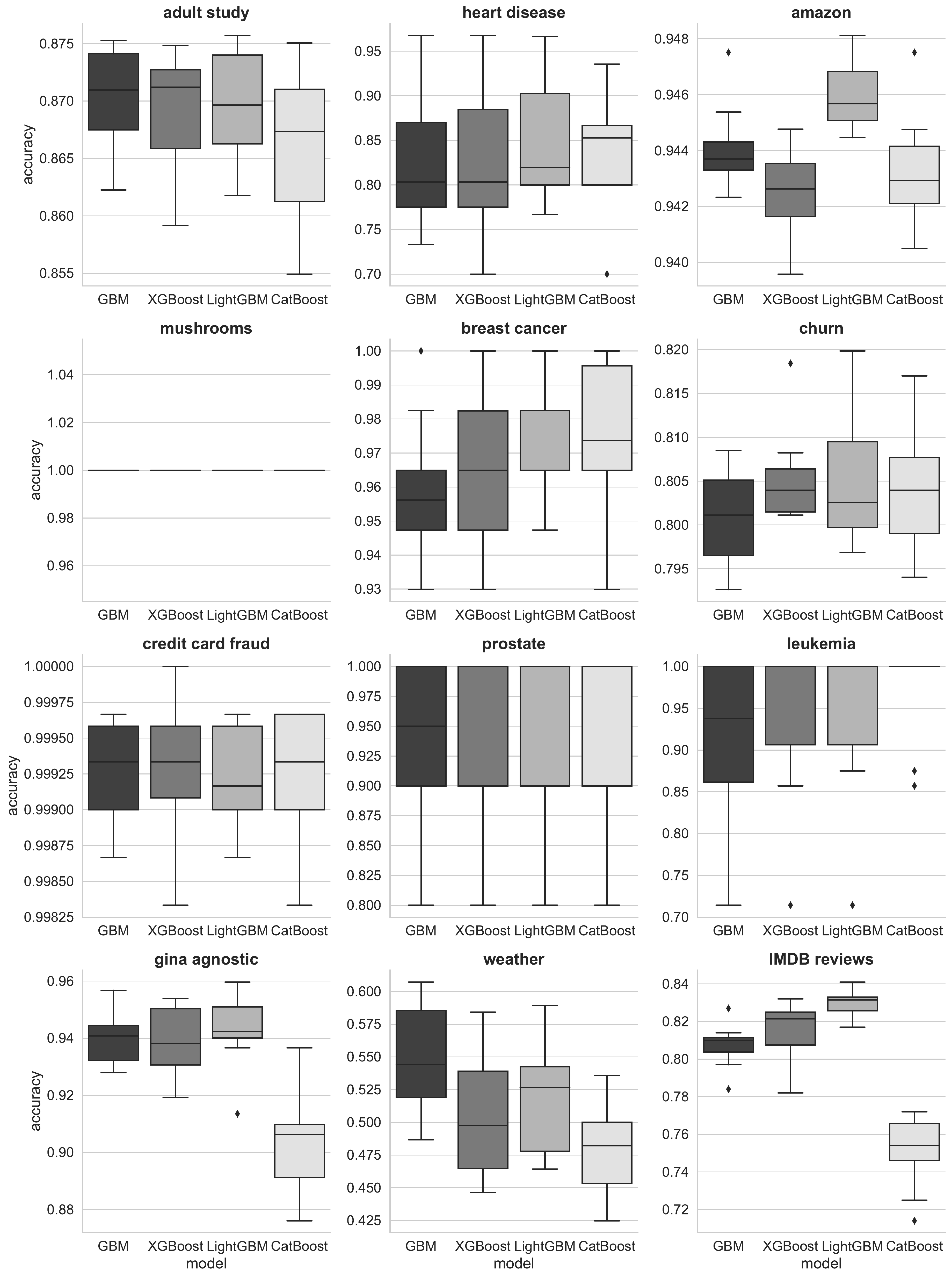}}
	\caption{Accuracy distributions across 12 datasets with randomized search}
	\label{fig:randomized_accuracy}
\end{figure}

\begin{figure}[H]
	\centering
		\scalebox{0.43}{\includegraphics{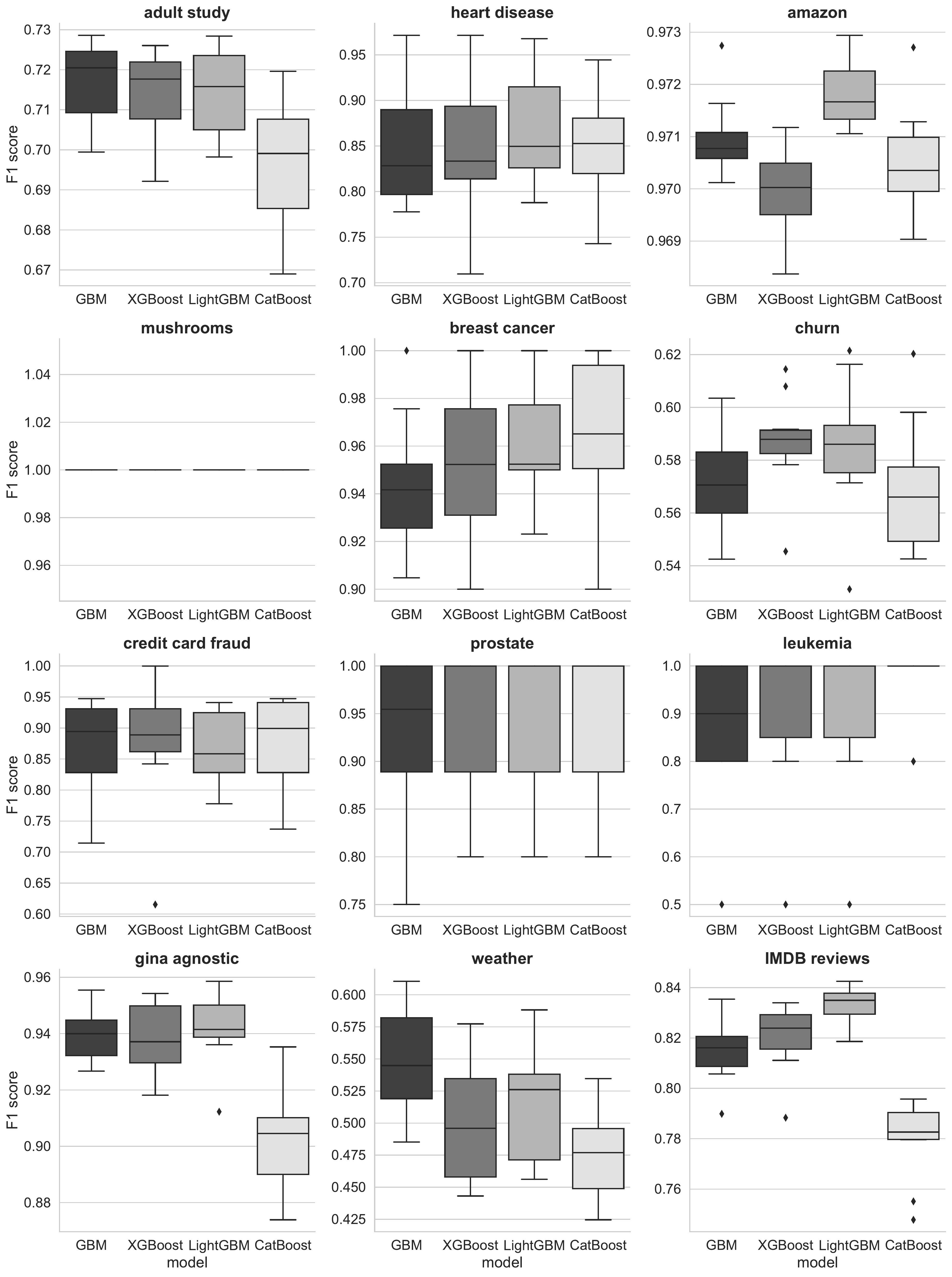}}
	\caption{F1 score distributions across 12 datasets with randomized search}
	\label{fig:randomized_f1}
\end{figure}

\begin{figure}[H]
	\centering
		\scalebox{0.43}{\includegraphics{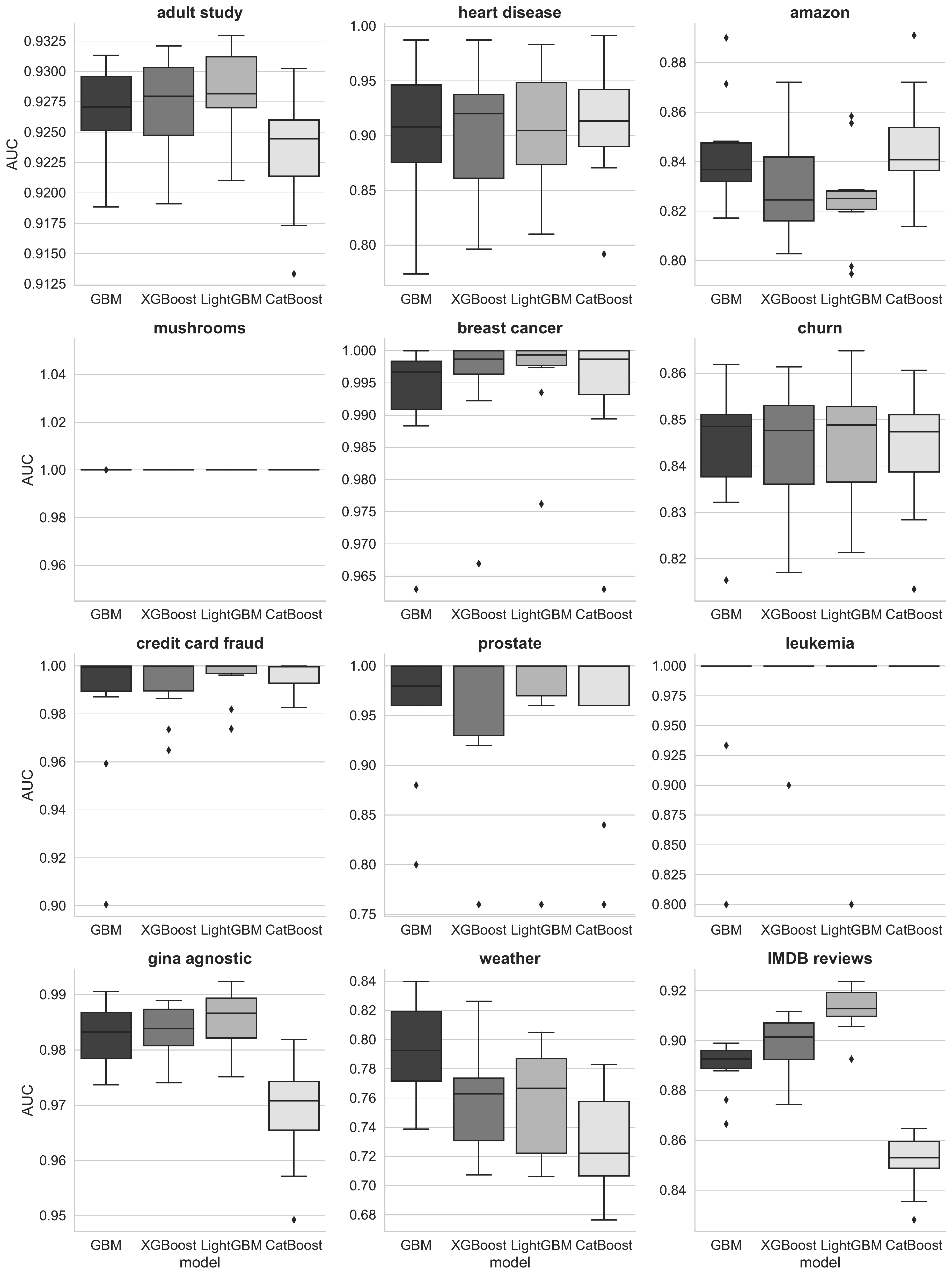}}
	\caption{AUC score distributions across 12 datasets with randomized search}
	\label{fig:randomized_AUC}
\end{figure}

\begin{figure}[H]
	\centering
		\scalebox{0.43}{\includegraphics{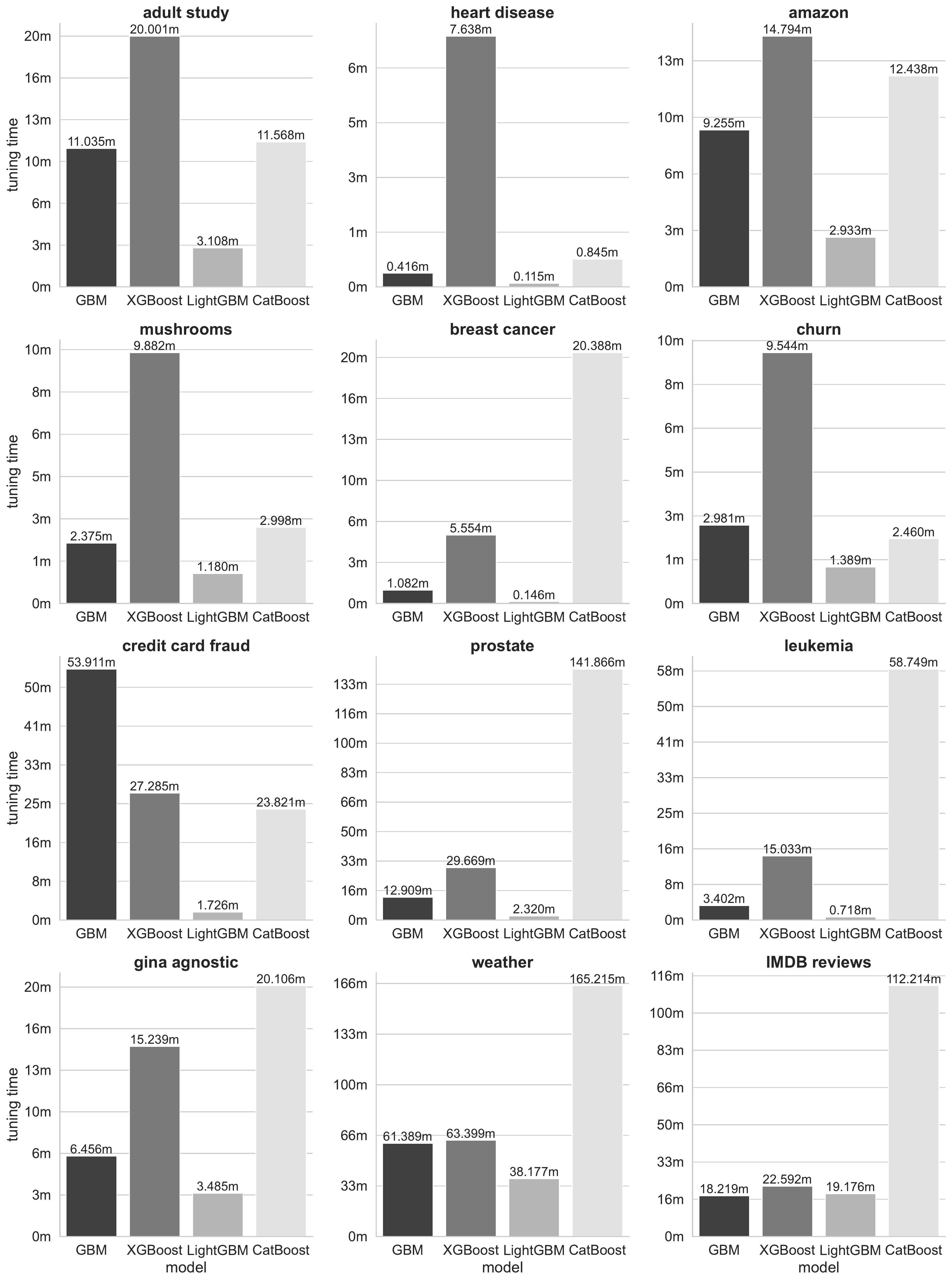}}
	\caption{Tuning times of the models across 12 datasets with randomized search}
	\label{fig:randomized_tuning_times}
\end{figure}

\begin{figure}[H]
	\centering
		\scalebox{0.43}{\includegraphics{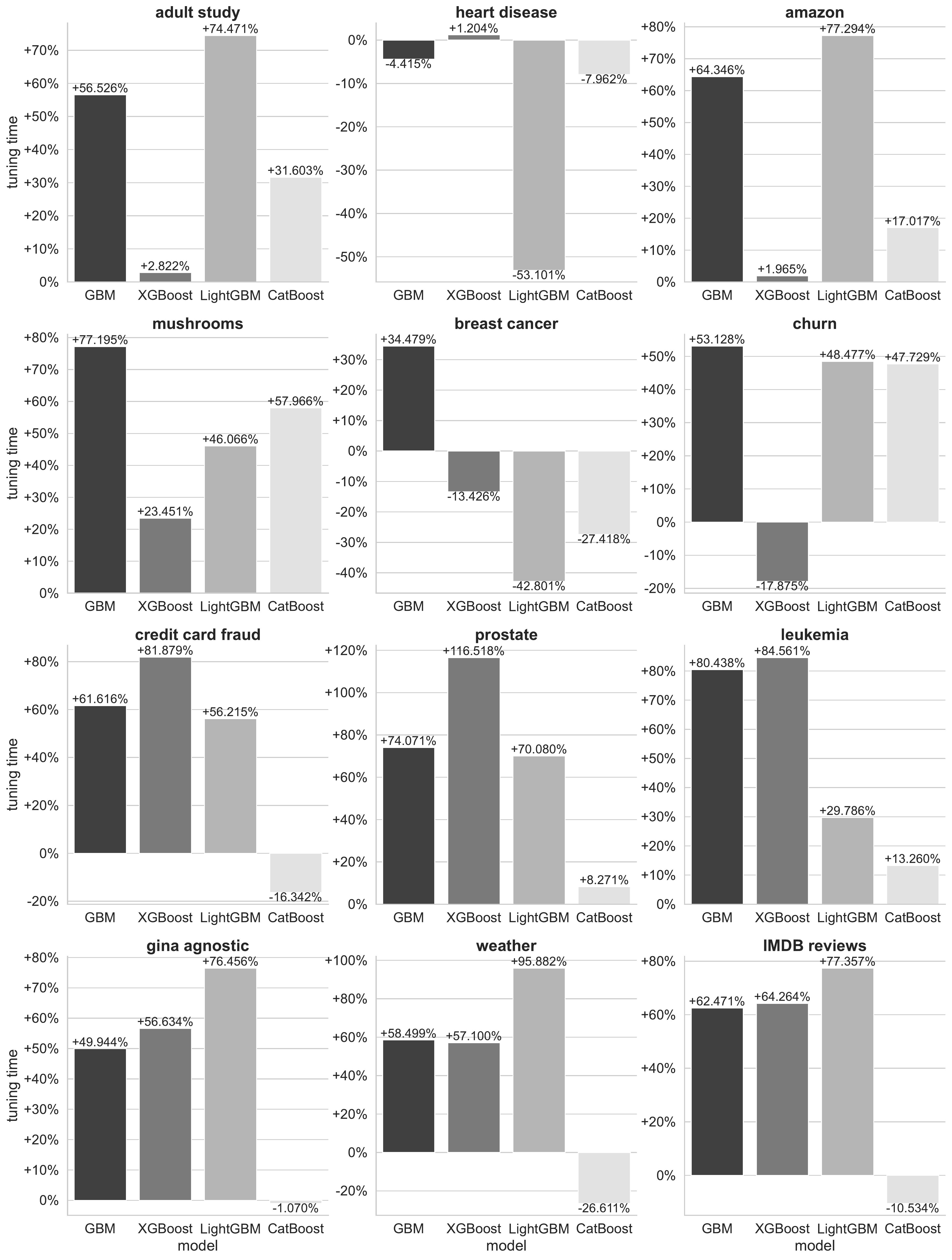}}
	\caption{Differences between tuning times using randomized search and Bayesian optimization}
	\label{fig:diffs}
\end{figure}